%% file: 00_MAIN.tex
\pdfoutput=1
\PassOptionsToPackage{table}{xcolor}
\documentclass[11pt]{article}

\usepackage[]{acl}

\usepackage{times}
\usepackage{latexsym}

\usepackage[T1]{fontenc}

\usepackage[utf8]{inputenc}

\usepackage{microtype}
\usepackage{caption,subcaption}
\usepackage{tcolorbox}
\usepackage{multirow}
\usepackage{float}
\usepackage{tabularx}
\usepackage{arydshln}
\usepackage{booktabs}
\newcommand{\rparagraph}[1]{\vspace{1.2mm}\noindent\textbf{#1.}}
\newcommand{\sparagraph}[1]{\vspace{0.0mm}\noindent\textbf{#1.}}

\usepackage[inline]{enumitem}
\usepackage{amsmath}
\usepackage[table]{xcolor}
\usepackage{tikz} 
\usepackage{pgfplots}

\definecolor{Gray}{gray}{0.92}
\definecolor{LightGray}{gray}{0.96}
\definecolor{LightCyan}{rgb}{0.92,0.968,0.968}
\definecolor{myyellow}{HTML}{db3e00}
\definecolor{myblue}{HTML}{004dcf}
\definecolor{myred}{HTML}{b80000}
\definecolor{mygreen}{HTML}{008b02}
\newcolumntype{Y}{>{\centering\arraybackslash}X}

\title{Reranking \textit{Over}generated Responses for \\ End-to-End Task-Oriented Dialogue Systems}

\author{
Songbo Hu, Ivan Vuli\'{c}, Fangyu Liu, Anna Korhonen \\
Language Technology Lab, University of Cambridge \\
  \texttt{\{sh2091,iv250,fl399,alk23\}@cam.ac.uk}
}

\begin{document}
\maketitle
\begin{abstract}

End-to-end (E2E) task-oriented dialogue (ToD) systems are prone to fall into the so-called `likelihood trap', resulting in generated responses which are dull, repetitive, and often inconsistent with dialogue history. Comparing ranked lists of multiple generated responses against the `gold response' (from evaluation data) reveals a wide diversity in response quality, with many good responses placed lower in the ranked list. The main challenge, addressed in this work, is then how to reach beyond \textit{greedily generated} system responses, that is, how to obtain and select such high-quality responses from the list of \textit{overgenerated} responses at inference \textit{without availability} of the gold response. To this end, we propose a simple yet effective reranking method which aims to select high-quality items from the lists of responses initially overgenerated by the system. The idea is to use any sequence-level (similarity) scoring function to divide the semantic space of responses into high-scoring versus low-scoring partitions. At training, the high-scoring partition comprises all generated responses whose similarity to the gold response is higher than the similarity of the greedy response to the gold response. At inference, the aim is to estimate the probability that each overgenerated response belongs to the high-scoring partition, given only previous dialogue history. We validate the robustness and versatility of our proposed method on the standard MultiWOZ dataset: our methods improve a state-of-the-art E2E ToD system by 2.0 BLEU, 1.6 ROUGE, and 1.3 METEOR scores, achieving new peak results. Additional experiments on the BiTOD dataset and human evaluation further ascertain the generalisability and effectiveness of the proposed framework.

\end{abstract}

\section{Introduction}
\label{s:introduction}

\input{01_introduction}

\section{Related Work}
\label{s:rw}
\input{02_rw}

\section{Post-Generation Response Reranking}
\label{s:methodology}
\input{03_methodology}

\section{Experimental Setup}
\label{sec:setup}
\input{04_experimental}

\section{Results and Discussion}
\label{s:results}
\input{05_results}

\section{Conclusion}
\label{s:conclusion}
\input{06_conclusion}

\section*{Acknowledgements}
This work has been supported by a Huawei research donation to the Language Technology Lab. The work of IV has been supported by a personal Royal Society University Research Fellowship (no 221137; 2022-2027). SH has been supported by Cambridge International Scholarship. FL has been supported by Grace \& Thomas C. H. Chan Cambridge International Scholarship. We thank our human participants for helping with evaluation.

\section{Limitations}
\label{s:limitations}
\input{07_limitations}

\bibliography{anthology,custom}
\bibliographystyle{acl_natbib}

\clearpage
\appendix
\input{xx_appendix}

\end{document}

%% file: 01_introduction.tex
Task-oriented dialogue (ToD) systems~\cite{williams2007partially, young2013pomdp} have received increasingly intensified research interest as they can assist humans with or automate many tasks effectively, this way contributing to technology expansion and inclusion \cite{Raux:2003,ElAsri:2017sigdial,budzianowski2018multiwoz,Laranjo:2018}. The natural language generation (NLG) module, also dubbed \textit{response generation}, is a critical component of any ToD system. Besides the necessary requirement to maintain semantic coherence during conversation, NLG also impacts user experience and satisfaction with a system.

Enabled by the recent advances in pretrained language models (PLMs)~\cite{radford2019language, raffel2020exploring}, now a \textit{de facto} approach to NLG is fine-tuning autoregressive language models on a domain-specific dialogue dataset~\cite{lin2020mintl, peng2022godel}. However, this approach still suffers from several crucial issues. \textbf{1)} Standard autoregressive models over-rely on local context~\cite{khandelwal2018sharp, sun2021long}, whereas many desirable properties of a dialogue response, such as consistency or coherence, can be captured only when taking into account dialogue history \cite{Zaib:2021survey}. \textbf{2)} Autoregressive LMs make predictions conditioned on the ground truth during training but on their own predictions during decoding, creating a disparity known as `exposure bias'~\cite{bengio2015scheduled, ranzato2015sequence, du2019empirical}. \textbf{3)} Finally, decoding dialogue responses from PLMs can easily fall into the so-called `likelihood trap' ~\cite{see-etal-2019-massively, zhang-etal-2021-trading}; here, high-likelihood (i.e., low-perplexity) sequences produced by greedy decoding or beam search tend to be dull and repetitive \cite{see-etal-2019-makes}. Truncated sampling methods, such as top-k~\cite{fan-etal-2018-hierarchical}, nucleus~\cite{Holtzman2020The}, and typical sampling~\cite{meister2022typical} also tend to produce text with inconsistencies, hallucinations, factual errors, or commonsense issues~\cite{massarelli-etal-2020-decoding, dou-etal-2022-gpt, krishna-etal-2021-hurdles,dziri-etal-2022-origin}.

\begin{figure}[!t]
    \centering
    \includegraphics[width=0.89\linewidth]{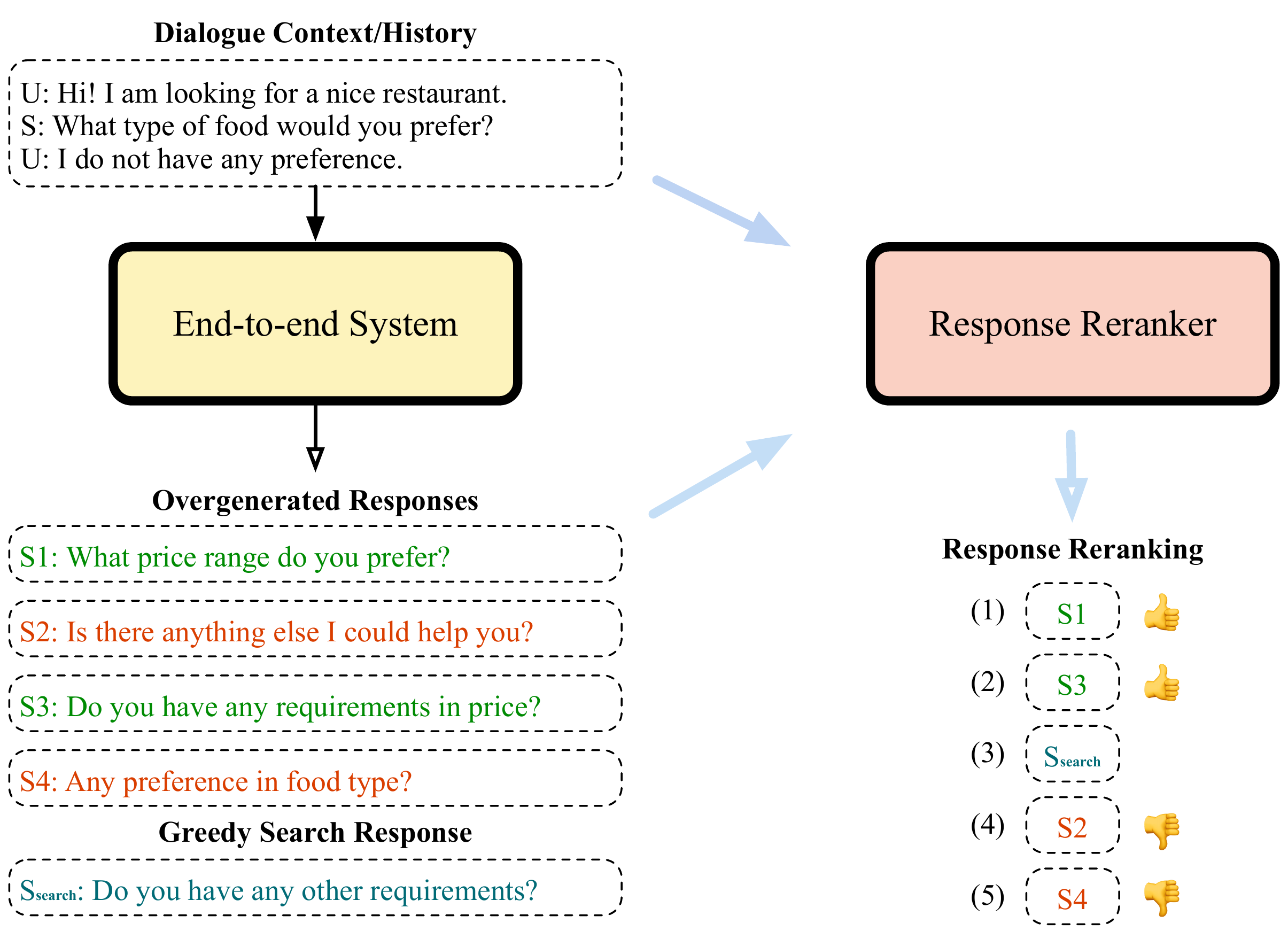}
    \caption{An illustration of our proposed reranking method. S: System; U: User. A reranking model is trained to rank a set of overgenerated responses from an end-to-end ToD system solely based on dialogue context/history. According to a predefined scoring function (e.g., cosine similarity, BLEU), a good candidate should have a score higher than that of the greedy search response. After reranking, the output of the base ToD E2E system is steered towards higher-quality responses.}
    \vspace{-1mm}
    \label{fig:fig1}
    \vspace{-2mm}
\end{figure}

To tackle these issues, we propose a \textit{post-generation reranking method for ToD}. The focus is on end-to-end (E2E) ToD systems, where NLG is modelled as a sequence-to-sequence problem. In particular, an E2E ToD system utilises a neural model such as T5~\cite{raffel2020exploring} or BART~\cite{lewis2019bart} to generate a surface form response conditioned on dialogue history and other context (e.g. dialogue domain ontology).

The method, illustrated in Figure~\ref{fig:fig1}, reranks a set of responses generated by any E2E ToD system solely based on the preceding dialogue context. Inspired by previous work on conversational representation learning~\cite[\textit{inter alia}]{mehri-etal-2019-pretraining, Humeau2020Poly-encoders, vulic2021convfit}, we fine-tune any input PLM (e.g., BERT, RoBERTa) into a conversational encoder which learns fine-grained interactions between target responses and dialogue context.\footnote{Unlike prior work, we do not use any task annotation directly for fine-tuning (e.g., prior work relied on intent labels to specialise PLMs towards particular intent detection tasks).} The proposed method is designed as a two-stage approach, see Figure~\ref{fig:fig2} later. In \textbf{Stage 1} we adaptively fine-tune the input PLM according to a specified \textit{scoring function} (e.g., cosine similarity, BLEU) and then use it to divide the semantic space (i.e., corresponding sets of generated responses) into \textit{high-scoring} and \textit{low-scoring} partitions based on their similarity to the gold response (according to the scoring function). Subsequently, in \textbf{Stage 2} such a specialised dialogue encoder allows reranking the generated responses based on discriminative classification or similarity-based retrieval, without leveraging the gold response. In turn, this enables us to run reranking without gold responses at inference.

In our main experiments on the standard MultiWOZ 2.0 dataset~\cite{budzianowski2018multiwoz}, we run the proposed method on top of the state-of-the-art (SotA) MinTL E2E ToD system~\cite{lin2020mintl}. Our method achieves substantial gains across the board and with different underlying PLMs: relying on cosine similarity as the scoring fuction, our rerankers achieve new SotA results of 20.0 ($\uparrow$2.0) BLEU, 32.8 ($\uparrow$1.6) ROUGE, and 36.9 ($\uparrow$1.3) METEOR on MultiWOZ. Furthermore, using the actual evaluation metric also as the scoring criterion in Stage 2, we further push performance to 20.3 ($\uparrow$2.3) BLEU, 33.6 ($\uparrow$2.4) ROUGE, and 40.0 ($\uparrow$4.4) METEOR. Ablation studies, additional experiments on the English portion of the BiToD dataset~\cite{lin2021bitod}, and human-based evaluation further verify the usefulness of the proposed method.\footnote{The code is available online at: \url{https://github.com/cambridgeltl/response_reranking}.}

%% file: 02_rw.tex
\sparagraph{NLG in ToD Systems}
In modular ToD systems \cite{williams2007partially}, the NLG task is defined as the surface realisation from dialogue acts, i.e., task-specific meaning representations on which the system operates.\footnote{An example dialogue act looks like this: \textit{Restaurant-Inform: [[bookpeople, 5], [booktime, 19:45]]} with its surface form utterance \textit{There will be 5 of us and 19:45 would be great.}} A plethora of autoregressive NLG models were proposed based on RNNs~\cite{wen2015stochastic, wen-etal-2015-semantically, tran-etal-2017-neural, tran-nguyen-2017-natural} and Transformers~\cite{peng2020few,Zhang:2022arxiv}. 
E2E ToD systems have received increasing attention recently~\cite{wen2016network, bordes2017learning, lei2018sequicity, eric-manning-2017-copy, eric-etal-2017-key, lin2020mintl, he2022galaxy}. They treat dialogue as a sequence transduction problem from dialogue context and other context (e.g., dialogue domain ontology) to a target system response. We focus on improving NLG performance of E2E ToD systems, proposing a post-generation reranking method which operates on the E2E system's outputs. %

\rparagraph{Post-Generation Reranking}
It has been well-studied in the Machine Translation (MT) community. Noisy Channel Modeling~\cite{ng2019facebook, yee-etal-2019-simple} is a widely used reranking scheme for NMT, parameterising the noisy channel probability with a seq2seq model. A reranker has also been implemented with an RNN language model~\cite{gulcehre2017integrating}, an energy-based model~\cite{bhattacharyya2020energy}, and masked language models~\cite{salazar2019masked, liu2021simcls}.

In dialogue, reranking methods were mostly investigated for open-ended systems, aiming to increase their response diversity~\cite{sordoni-etal-2015-neural, li-etal-2016-diversity, shao-etal-2017-generating}, to improve conversational `engagingness' by integrating human feedback~\cite{gao-etal-2020-dialogue}, and to enhance fluency and semantic correctness~\cite{baheti-etal-2020-fluent}. However, post-generation reranking has not been as widely explored in the ToD context. Compared to open-ended systems, generated outputs from ToD systems are highly semantically similar, adding a substantial challenge to reranking, reaching beyond simple topic shifts of the responses. Notably, rerankers based on convolution~\cite{wen2015stochastic}, RNN~\cite{dusek-jurcicek-2016-sequence}, RoBERTa~\cite{harkous-etal-2020-text}, and cross-attention~\cite{juraska-walker-2021-attention} were proposed, which all crucially assume access to the ground truth dialogue act representation; we do not assume its availability in our more difficult and realistic setup.

\rparagraph{Response Selection for ToD}
Another line of research focuses on retrieval-based methods for providing system responses in ToD. These methods retrieve a set of response candidates and subsequently select the most likely one (according to some matching function) as a final response \cite{ritter-etal-2011-data}. Different matching models have been proposed to measure the matching degree between a dialogue context and a response candidate, and rank the candidates accordingly~\cite[\textit{among others}]{wu-etal-2017-sequential, zhou-etal-2018-multi, weston-etal-2018-retrieve, lu-etal-2019-constructing, 10.1145/3357384.3358140, su2020dialogue, henderson-etal-2020-convert, Humeau2020Poly-encoders}. Unlike prior work, which typically ranks a set of predefined system response candidates, our post-generation reranking method combines generation-based and retrieval-based methods. Moreover, while previous work~\cite[\textit{e.g.}][]{weston-etal-2018-retrieve, dinan2018wizard, kim2020domain} augmented the dialogue context with retrieved knowledge \textit{before} generation, our method is a post-generation reranking method. In particular, our rerankers operate on a set of over-generated (and thus semantically close) responses; we thus need to capture very subtle nuances between different response candidates.

\rparagraph{Contrastive Learning for NLG}
Contrastive learning (CL)~\cite{chopra2005learning, schroff2015facenet, chen2020simple, he2020momentum} has been widely used in NLP for word-level~\cite{mikolov2013efficient, vulic-etal-2021-lexfit, liu-etal-2021-mirrorwic} and sentence representation learning~\cite{reimers2019sentence, wu2020clear, meng2021coco, liu2021fast, gao2021simcse}. Beyond representation learning, other work applies CL to open-ended text generation~\cite{krishna2022rankgen, su2022contrastive}. However, as posited by~\citet{krishna2022rankgen}, such a method may not be directly applicable to other generation tasks with a more constrained output space (e.g. NLG for ToD). For constrained generation tasks, \citet{liu2021simcls} apply CL to 
post-generation reranking for abstractive summarisation, and \citet{an2022cont} use CL for five generation tasks, but none of them relates to dialogue. 

Our paper applies contrastive learning within the context of NLG for ToD. In comparison to other NLG tasks, this context provides unique challenges. The context-response pairs within a ToD dataset usually violate the \textit{i.i.d.} assumption. In particular, randomly sampled negative utterances within a dataset might easily constitute, \textit{ipso facto}, a valid positive sample because of the one-to-many nature of ToD~\cite{zhao2017learning}. Moreove, multiple dialogues can describe very similar scenarios, exacerbating this issue. Our proposed method evades this issue via a novel \textit{heuristic thresholding procedure} (see \S\ref{s:methodology} later), removing the dependencies between contrastive pairs (a positive example for CL is not conditioned on a dialogue context).\footnote{In addition, previous work in non-ToD contexts~\cite{krishna2022rankgen} only utilises self-generated sentences to construct negative examples for CL. Our proposed method explores constructing both positive and negative pairs from model predictions. Foreshadowing, our results in \S\ref{s:results} demonstrate that this is crucial to enable effective reranking.}

%% file: 03_methodology.tex
\sparagraph{Motivation: An Oracle Experiment}
In an `oracle' experiment, where we assumed the availability of the ground truth response, we reranked the set of 20 `oversampled' responses from an E2E ToD system, using their sentence-level BLEU-based similarity to the ground truth. This reranking procedure yielded gains of 16.2 BLEU points, corroborating a similar finding from the MT literature \cite{bhattacharyya2020energy}. %
 However, the crucial issue is that at real-world `non-oracle' inference we do not have access to such ground truth responses. The observation from the oracle experiment indicates that: (i) there is ample room for improvement in NLG via reranking, while (ii) we need to disentangle the critical dependency on the ground truth response from the reranking process.

\rparagraph{Response Reranking: Preliminaries}
The task is similar to response selection \cite[\textit{among others}]{wu-etal-2017-sequential,henderson-etal-2019-training,Humeau2020Poly-encoders}. Given a dataset with $n$ examples $\mathcal{D} = \{\mathcal{D}^{(1)}, \mathcal{D}^{(2)},\cdots,\mathcal{D}^{(n)}\}$, each example $\mathcal{D}^{(i)}\in\mathcal{D}$ contains the pair $({c}^{(i)}, {r}^{(i)})$, where ${c}^{(i)}$ is the dialogue context and ${r}^{(i)}$ is the response; $\mathbf{c}^{(i)}$ and $\mathbf{r}^{(i)}$ denote their respective representations/embeddings. During training, the task is to learn a scoring function $s(\cdot, \cdot)$ that assigns a matching score for any context--response pair. At inference, response reranking involves a dialogue model  $P_{\mathrm{MLE}}({r} \mid {c})$ and an evaluation metric $\operatorname{M}(\cdot, \cdot)$. Given the context $c$, we sample a set of responses ${R} = \{{r}_{1},{r}_{2}\ldots{r}_{j}\}$ from $P_{\mathrm{MLE}}(r \mid c)$. The task of a response reranker is to assign a score $s(\cdot, \cdot)$ for each context--response pair and select a response based on this score, e.g. $\operatorname{argmax}_{{r} \in {R}} s\left(\mathbf{c}, \mathbf{r}\right)$. Response reranking is tasked to improve the evaluation score $\operatorname{M}(c, r)$.

\begin{figure}[!t]
    \centering
    \includegraphics[width=0.92\linewidth]{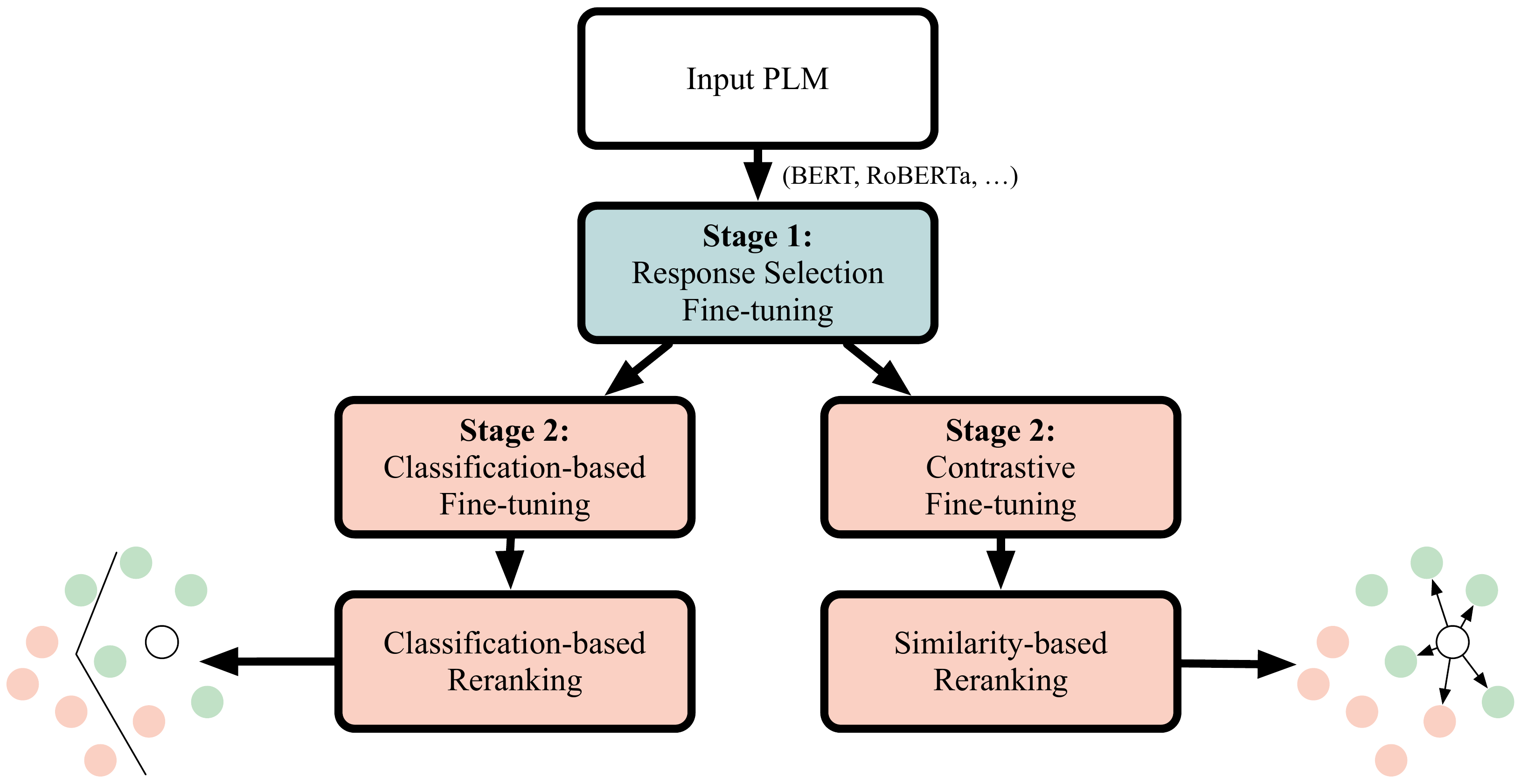}
    \vspace{-1.5mm}
    \caption{An overview of the two-stage reranking process. \textbf{Stage 1} fine-tunes any input PLM into a response selection model with in-domain data. \textbf{Stage 2} further fine-tunes the Stage 1 model into a response reranking model, more sensitive to fine-grained interaction between dialogue context and candidate responses.}
    \label{fig:fig2}
    \vspace{-2mm}
\end{figure}

\subsection{Methodology}
\label{sec:methodlogy}

An effective response reranker for ToD should capture \textit{subtle} differences among a set of \textit{highly similar} candidates generated by a fine-tuned E2E ToD model. As mentioned in \S\ref{s:rw}, this setup is considerably more difficult than selecting the best response from randomly sampled confounders from a dialogue dataset \cite{henderson-etal-2019-training,gunasekara-etal-2019-dstc7}. In a preliminary experiment, we followed the standard response selection setup~\cite{wu-etal-2017-sequential, zhou-etal-2018-multi, 10.1145/3357384.3358140} and aimed at distinguishing between the ground truth positive example from randomly sampled negatives. We found out that those baselines perform well on the response selection task but achieve near-random performance in our response reranking task focused on highly semantically similar candidate responses. Therefore, in what follows, we propose a novel fine-tuning framework to deal with the much more challenging (re)ranking scenario. %

\rparagraph{Method in a Nutshell}
We train a generative E2E dialogue model $P_{\mathrm{MLE}}({r}\mid{c})$ on a dialogue dataset $\mathcal{D}$. Subsequently, for each training example $({c}, {r})$ in the training set, we sample a set of responses $\mathcal{R} = \{{r}_{1},{r}_{2}\ldots{r}_{j}\}$ from $P_{\mathrm{MLE}}({r}\mid{c})$, where $j$ denotes the number of over-generated responses. For each ${r}_{k} \in \mathcal{R}$, we calculate its score based on a \textit{scoring function} ${s}_{k} = s(\mathbf{r}_{k} , \mathbf{r})$, where $\mathbf{r}$ is the representation of the ground truth response. Unless stated otherwise, the default scoring function is defined as the cosine similarity based on a general-purpose sentence encoder.\footnote{We use the \texttt{all-mpnet-v2}~\cite{reimers2019sentence} as a robust, efficient and high-performing choice.} We then cluster the sampled responses $\mathcal{R}$, based on their respective scores and a defined \textit{thresholding procedure} (see \S\ref{ss:stage1}), into a high-scoring set $\mathcal{R}_{high}$ and a low-scoring set $\mathcal{R}_{low}$. During training, the reranking model aims to directly capture the distinction between $\mathcal{R}_{high}$ and  $\mathcal{R}_{low}$. At inference, the reranking model scores and ranks a candidate response ${r}_{k}$ based on the \textit{probability that the generated response is drawn from the high-scoring set}, namely $P({r}_{k} \in \mathcal{R}_{high}\mid{c})$. Again, we stress that our reranking model does not require access to the ground truth during inference. Instead, the conditional probability serves as a `proxy' function, indicating the likelihood that the candidate is indeed a valid response. This way, we make a transition from `oracle'-based training with ground truth to the `non-oracle' inference.

Following~\citet{vulic2021convfit}, we propose a two-stage fine-tuning procedure, with two types of reranking models in the second stage: a \textit{classification-based} model and a \textit{similarity-based} model, as illustrated in Figure~\ref{fig:fig2}. The framework can be applied on top of any input (Transformer-based) encoder $\mathbf{e} = enc_{\theta }(t)$ parameterised by $\theta$, which encodes textual input ${t}$ into a sentence embedding. Unless stated otherwise, we use BERT(-base) \cite{devlin-etal-2019-bert} as our default encoder.

\subsection{Stage 1: Response Selection}
\label{ss:stage1}

In Stage 1, we conduct adaptive fine-tuning in the response selection task \cite{vulic2021convfit}, which transforms the input PLM into a text encoder that is better aligned with the end-task \cite{Ruder:2021blog} of response reranking. We rely on a standard cross-encoder architecture that directly models the interaction between the context and the candidate responses. Each data example is a tuple $(c, {r},{l})$, where ${l} \in\{0,1\}$ is a binary label indicating if $r$ is the ground truth response to ${c}$. In fact, for each dialogue $({c}^{(i)}, {r}^{(i)}) \in \mathcal{D}$ , we construct a positive example $({c}^{(i)}, {r}^{(i)}, 1)$. We then randomly sample a set of $N_{r}$ negative responses $\mathcal{R}_{i,-}$ per each positive response $r^{(i)}$ from other tuples following prior work on response selection: for each ${r}^{(j)} \in \mathcal{R}_{i,-}$ it holds $i \neq j$, and we construct final negative samples as follows: $({c}^{(i)}, {r}^{(j)}, 0)$.

The goal of Stage 1 is to fine-tune the input PLM/encoder into a statistical model parameterised by ${\theta}$ to compute $P_{\theta }({l} | {c},{r})$. Given a training example $({c}, {r}, {l})$, the model is trained to predict the correct label by encoding the concatenation of a context response pair $[{c},{r}]$. To this end, the representation of the ``\texttt{[CLS]}'' token is subsequently projected down to two logits and passed through a softmax layer to form a Bernoulli distribution indicating the positive (1) or the negative (0) label.

\subsection{Stage 2: Response Reranking}
\label{sec:stage2}
Each data entry for response reranking in Stage 2 is again a tuple $({c}, {r}, {l})$, where ${l} \in\{0,1\}$ is a binary label. We construct those data entries as follows. First, for each dialogue item $({c}, {r}) \in \mathcal{D}$, we generate a set of responses $\mathcal{R} = \{{r}_{1},{r}_{2}\ldots{r}_{j}\}$ and a greedy search response ${r}_{search}$. We then calculate a pair-wise score ${s}_{k}$ between each generated response ${r}_{k} \in \mathcal{R}$ and the ground truth response ${r}$, relying on some \textit{scoring function} (e.g., cosine similarity between their sentence embeddings). Similarly, we calculate a score ${s}_{search}$ for the greedy search response ${r}_{search}$. The score ${s}_{search}$ is used as a local \textit{threshold value} that splits the set of generated responses into positive (i.e., `high-quality') and negative (`low-quality') responses as follows.

If ${s}_{k}  \geq {s}_{search}$, we add the generated response ${r}_{k}$ to the high-quality set $\mathcal{R}_{high}$; if ${s}_{k}  < {s}_{search}$, we add ${r}_{k}$ to the set $\mathcal{R}_{low}$.\footnote{We note that there are other options to split the candidate responses into high-scoring and low-scoring sets $\mathcal{R}_{high}$ and $\mathcal{R}_{low}$ (e.g., selecting the top N\% responses for $\mathcal{R}_{high}$). However, our proposed method is hyperparameter-free and is also conditioned on the score of the greedy response, where greedy search is the default decoding strategy of many standard E2E ToD systems \cite{lin2020mintl}.} Since the cardinality of the two sets may differ, we downsample the larger set to the size of the smaller one: $min(\lvert \mathcal{R}_{high} \rvert , \lvert \mathcal{R}_{low} \rvert)$. Following that, for each ${r}_{k} \in \mathcal{R}_{high}$, we construct a positive example for fine-tuning $({c}, {r}_{k}, 1)$; for each ${r}_{k} \in \mathcal{R}_{low}$, we construct a negative example $({c}, {r}_{k}, 0)$. We construct such examples from the entire training set, and they are then used for two types of reranking: \textit{classification-based} and \textit{similarity-based}, described in what follows, with additional illustrations in Figure~\ref{fig:fig3} in Appendix~\ref{app:stage2_fig}.

\rparagraph{Classification-Based Reranking}
\label{sec:classification}
This procedure is identical to Stage 1. However, the reranking models now learn to rerank overgenerated (and semantically similar) responses according to positive and negative examples corresponding to respective sets $\mathcal{R}_{high}$ and $\mathcal{R}_{low}$. Given a training example $({c}, {r}, {l})$, the encoder first encodes the $[{c},{r}]$ and the contextualised representation of the ``\texttt{[CLS]}'' token is subsequently used to compute $P_{\theta }({l}|{c},{r})$ as in Stage 1 using the standard cross-encoder architecture~\cite[\textit{e.g.}][]{wolf2019transfertransfo, urbanek2019learning}.\footnote{For response reranking, the candidate set is usually dynamic (i.e., generated on-the-fly) and comprises 10-20 items. In such setups, using cross-encoders is computationally feasible \cite{urbanek2019learning}.} At inference, given $\mathbf{c}$ and a set of generated responses $\mathcal{R}$, we rerank and select the final response based on its score: $\operatorname{argmax}_{\mathbf{r} \in \mathcal{R}} P({l} = 1 | {c},{r})$.

\rparagraph{Similarity-Based Reranking}
\label{sec:similarity}
Similarity-based classification has demonstrated promising results in intent detection for ToD~\cite{zhang-etal-2020-discriminative, vulic2021convfit} and other NLP tasks~\cite{sarwar2022neighborhood, kassner-schutze-2020-bert}, particularly when data is scarce. We thus also propose a similarity-based reranker in Stage 2, based on contrastive fine-tuning and KNN retrieval.

The aim is to fine-tune the input encoder so that it encodes all context-response pairs from $\mathcal{R}_{high}$ into coherent clusters, clearly separated from low-scoring pairs from $\mathcal{R}_{low}$. Here, we utilise the label ${l}$ during training only implicitly, allowing us to formulate reranking as a sentence similarity task. In particular, for a training example $({c}, {r}, {l})$, the encoder first encodes $[{c},{r}]$, where the encoding $\mathbf{e} = enc_{\theta }([{c},{r}])$ is created via mean-pooling over the constituent subwords' embeddings.

We randomly divide the training examples into mini-batches, and fine-tune the input encoder via the standard Triplet Loss~\cite{schroff2015facenet}. For any pair of examples within a batch $(enc_{\theta }([{c}^{(i)}, {r}^{(i)}]),  {l}^{(i)})$ and $(enc_{\theta }([{c}^{(j)}, {r}^{(j)}]),  {l}^{(j)})$, the encoder parameters $\theta$ are optimised (i) to reduce the cosine distance between encodings of the pairs with the same label ${l}^{(i)} = {l}^{(j)}$, and (ii) to increase the distance otherwise. 
After Stage 2 fine-tuning, response scoring in the specialised encoder space $enc_{S_{2}}$ is then performed via similarity-based KNN inference~\cite{zhang-etal-2020-discriminative,vulic2021convfit}, using a subset of training examples as \textit{anchors}. For all anchors $({c}^{(i)}, {r}^{(i)},{l}^{(i)})$, we compute their encodings $\mathbf{e}^{(i)} = enc_{S_{2}}([{c}^{(i)}, {r}^{(i)}])$ in advance. 
For any candidate-response pair,  we obtain its encoding $\mathbf{e} = enc_{S_{2}}([{c}, {r}])$. We retrieve a set of $k$ nearest anchors from the full set of anchors. The scoring function $s(\cdot, \cdot)$ is defined as the proportion within the $k$ nearest anchors with a positive label. We rank and select the final response as follows: $\operatorname{argmax}_{{r} \in \mathcal{R}} s({c},{r})$.

%% file: 04_experimental.tex
\sparagraph{Training and Evaluation Data}
Our main experiments focus on the standard multi-domain MultiWOZ ToD dataset~\cite{budzianowski2018multiwoz}: in particular on its 2.0 version.

\rparagraph{Baseline E2E System}
The underlying E2E ToD system is MinTL ~\cite{lin2020mintl}, as a publicly available SotA model. It jointly learns dialogue state tracking and response generation with pre-trained seq2seq models.\footnote{MinTL~\cite{lin2020mintl} was the SotA system for E2E NLG on the MultiWOZ leaderboard until the recently published GALAXY system~\cite{he2022galaxy} surpassed its performance by a 0.2 BLEU score. See the MultiWoZ leaderboard at \url{https://github.com/budzianowski/multiwoz}.} However, we note that the proposed reranking method can be applied to any E2E dialogue system with autoregressive response generation~\cite[\textit{e.g.,}][]{wen2016network,he2022galaxy}.

\rparagraph{Evaluation Metrics}
Following the standard MultiWOZ setup, we use the corpus BLEU score~\cite{papineni2002bleu} as our primary evaluation metric, and all the scores are computed with delexicalised utterances based on the DAMD system~\cite{DBLP:conf/aaai/ZhangOY20}.\footnote{In the case of delexicalised dialogues, all the slot values in the context and responses are replaced a predefined placeholder (e.g. \textit{[value\_name] is an [value\_price] [value\_food] restaurant on the [value\_area] . do you need to know more ?}).} We also report ROUGE-L~\cite{lin2004rouge} and METEOR~\cite{banerjee2005meteor} as two other standard NLG evaluation metrics.

\rparagraph{Input PLMs for Reranking}
Our method can be implemented with any Transformer-based~\cite{vaswani2017attention} PLM. To analyse the impact of the input PLM (see Figure~\ref{fig:fig2}), we experiment with several popular PLMs: BERT~\cite{devlin-etal-2019-bert}, RoBERTa~\cite{liu2019roberta}, and their distilled versions~\cite{sanh2019distilbert}. We additionally experiment with supervised sentence encoders: SimCSE~\cite{gao2021simcse} and other popular encoders from the sentence-transformers (i.e., SBERT) repository~\cite{reimers2019sentence}. Table~\ref{tab:input_plms} in the appendix lists all the input models we use along with their checkpoints from the HuggingFace repository~\cite{wolf2019huggingface}.

\rparagraph{Hyperparameters and Optimisation}
The default decoding strategy for MinTL is the greedy search. In our reranking experiments, unless stated otherwise, we over-generate 20 responses with nucleus sampling~\cite{Holtzman2020The} from the top-0.7 portion of the probability mass, a standard choice.

We implement all reranking models via the SBERT repository~\cite{reimers2019sentence}, which is built on top of the HuggingFace repository~\cite{wolf2019huggingface}. Table~\ref{tab:hyperparameters} in the appendix lists the search set of model hyperparameters (which differ from the default SBERT-suggested values), along with the finally set values. The grid search was conducted on the dev set, based on BLEU. 128 is the maximum batch size with BERT base and RoBERTa for Stage 2 fine-tuning. Following \citet{reimers2019sentence, vulic2021convfit}, we use the AdamW optimiser~\cite{loshchilov2017decoupled} in the default SBERT setting: the learning rate is $2$e-$5$; warmup of 0.1 and linear decay; the weight decay rate is 0.01. We rely on the triplet loss variant of \newcite{hermans2017defense}.\footnote{This is \texttt{BatchAllTripletLoss} in the SBERT repo; see the documentation and the paper for further details.}

For similarity-based reranking in Stage 2, we experiment with the anchor set size $k \in \{10, 100, 500, 1000, 5000\}$, randomly sampled from the full training data.

\subsection{Model Variants and Baselines}
\label{ss:variants}

We experiment with several model variants enabled by the proposed two-stage pipeline (see Figure~\ref{fig:fig2}):

\sparagraph{PLM+S1+S2}
This variant refers to the full pipeline, where PLM is any input PLM from Table \ref{tab:input_plms}. Stage 1 (S1) fine-tuning can be based on either lexicalised dialogues (\textit{S1:lex}) or delexicalised (\textit{S1:delex}) dialogues. After S1, we can further fine-tune the `S1' encoders via the classification-based or the similarity-based approach: \textit{S2:class} and \textit{S2:sim}. For instance, the configuration \textit{BERT+S1:delex+S2:sim} denotes the use of BERT as the input PLM, with delexicalised dialogues in Stage 1, and similarity-based Stage 2.

\sparagraph{PLM+S2}
This group is fine-tuned only relying on Stage 2 approaches, skipping Stage 1.

\sparagraph{PLM+S1}
This group is fine-tuned only for response selection with in-domain data, ignoring S2.

\sparagraph{PLM}
This variant refers to using out-of-the-box sentence encoders in the response reranking task. Since classification-based reranking requires a fine-tuned task-specific classification head, we only run our experiments with similarity-based reranking.

We also compare against two standard decoding strategies. \textbf{1) Greedy}. Greedy search has been widely used as the default decoding strategy NLG, also by the base MinTL system~\cite{lin2020mintl}.\footnote{Greedy search and beam search are used as the default decoding methods for many SotA E2E systems~\cite{lin2020mintl, lin2021bitod, he2022galaxy} as they typically outperform sampling algorithms in terms of BLEU.} \textbf{2) Sampling.} As mentioned, we apply nucleus sampling~\cite{Holtzman2020The} to sample responses from the top-0.7 portion of the probability mass. %

%% file: 05_results.tex
\begin{figure}[!t]
    \centering
    \includegraphics[trim={0 0 0 1.5cm}, width=0.87\columnwidth]{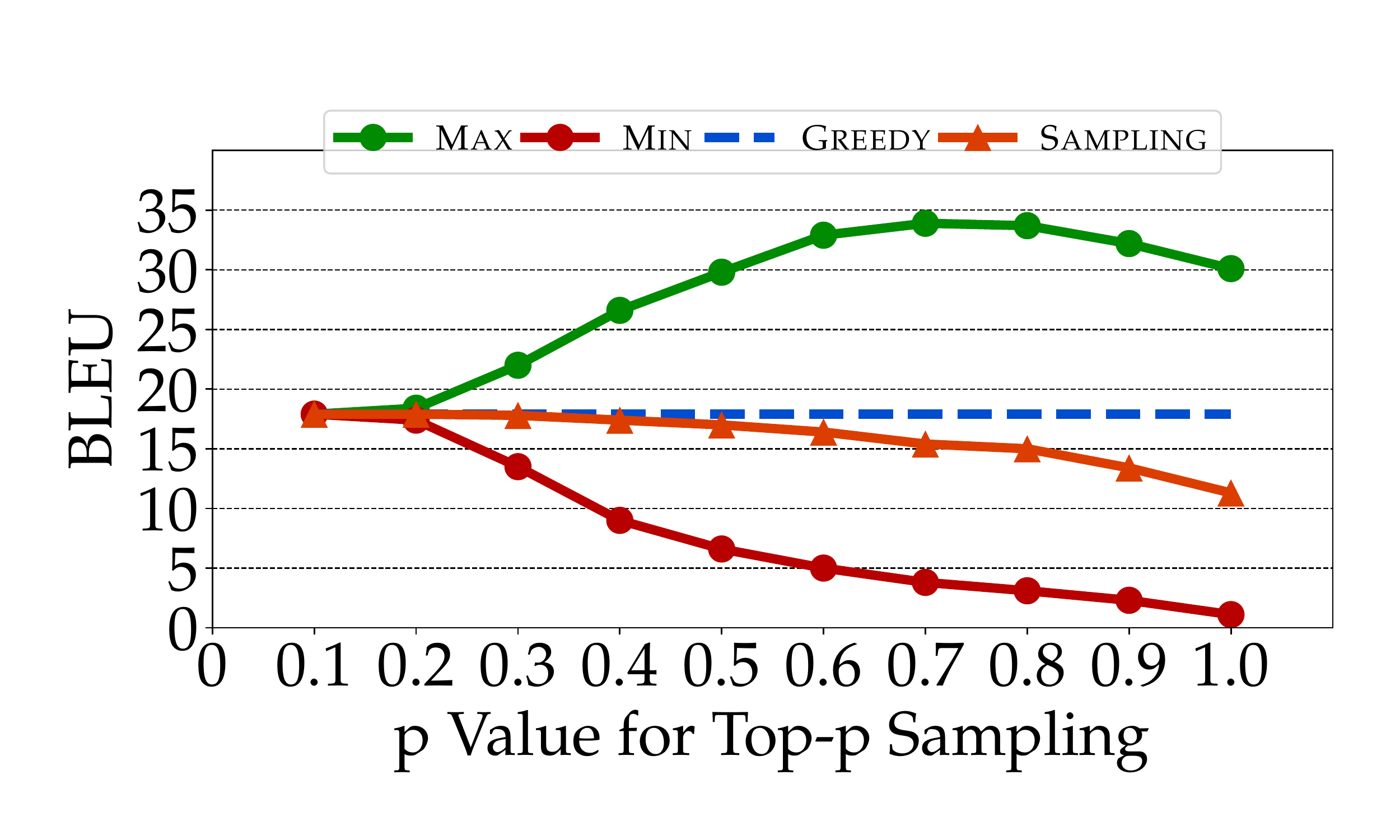}
    \vspace{-2.5mm}
    \caption{Corpus BLEU scores for MinTL on the MultiWoZ \textit{dev} set. Nucleus top-\textit{p} sampling with different \textit{p} values. \textcolor{mygreen}{Max} (\textcolor{myred}{Min}) performance is achieved by reranking 20 overgenerated samples based on their sentence BLEU score with the ground truth response and selecting the response with the maximum (or minimum) score. \textcolor{myblue}{Greedy} search is the standard decoding strategy.}
    \label{fig:find1}
    \vspace{-1mm}
\end{figure}

\sparagraph{(Preliminary) Motivational Empirical Evidence}
We first focused on examining the diversity of candidate responses generated by the underlying E2E ToD model, which depict the potential of reranking (see also \S\ref{s:methodology}). As revealed by Figure~\ref{fig:find1}, when overgenerating 20 responses with the nucleus sampling method~\cite{Holtzman2020The}, we can find responses in those sets that are of much higher-quality (as measured by BLEU) as well as of much lower-quality than the standard \textit{greedy} response. Upper bound improvements on MultiWoZ's dev set reach the peak with top-0.7 sampling: it is possible to improve BLEU up to 16.2 points if we always select the best response (according to BLEU) from the corresponding 20-response set. Simultaneously, while the diversity increases with the increase of $p$, the BLEU score for the \textit{Sampling} baseline (see \S\ref{ss:variants}) decreases. In sum, this empirical observation demonstrates the trade-off between greedy search algorithm and sampling algorithms for NLG in ToD, corroborates the dominance of the greedy search, and offers empirically driven motivation for our proposed reranking methods.

\begin{table}[!t]
\centering
\footnotesize
\def\arraystretch{0.9}
\begin{tabularx}{\linewidth}{l Y YYY}
\toprule
{} & \multicolumn{1}{c}{\bf Selection} & \multicolumn{3}{c}{\bf Reranking} \\
\cmidrule(lr){2-2} \cmidrule(lr){3-5} 
\bf {Variant} & {R@1} & {B} & {R} & {M} \\
\cmidrule(lr){2-2} \cmidrule(lr){3-5}
{Random Sampling} & {5.0} & {15.8} & {27.3} & {31.0} \\
{Greedy} & {--} & {18.0} & {31.2} & {35.6} \\
\hdashline
{BERT} & {--} & {17.0} & {29.4} & {33.6} \\
{SimCSE} & {--} & {16.7} & {29.0} & {33.2} \\
{all-mpnet} & {--} & {16.0} & {27.6} & {31.8} \\
\hdashline
{BERT+S1:delex} & {51.0} & {16.7} & {39.3} & {33.8} \\
{BERT+S1:lex} & {77.2} & {17.1} & {29.7} & {34.3} \\
{DRoB+S1:delex} & {48.0} & {16.6} & {29.0} & {33.4} \\
{DRoB+S1:lex} & {74.4} & {16.6} & {29.6} & {34.5} \\
\bottomrule
\end{tabularx}
\vspace{-1mm}
\caption{Performance of representative out-of-the-box sentence encoders and response selection models on the standard Response \textbf{Selection} task (R@1 = Recall@1), and on the final Response \textbf{Reranking} task, relying on the MultiWoZ test set. Similarity-based reranking without S2 fine-tuning is reported. B=BLEU;  R=ROUGE; M=METEOR; DRoB=DistilRoBERTa. Additional results (incl., classification-based reranking) with more input models are available in Table~\ref{tab:selection_old} in Appendix~\ref{app:multiwoz}.}
\label{tab:selection}
\vspace{-2mm}
\end{table}

\rparagraph{Performance of Standard Response Selection}
We then investigated the capability of standard response selection techniques from the literature (i.e., effectively running only S1 in our pipeline) to select the best response from the sets of overgenerated candidates. The results, summarised in Table~\ref{tab:selection}, reveal that response selection models and other out-of-box sentence encoders are not effective in the response reranking task. All the model variants score lower than the standard \textit{Greedy} search baseline in the reranking task despite the fact that S1 in-domain fine-tuning increases their response selection capabilities (e.g., compare their R@1 scores versus the random baseline). Moreover, the results indicate that selecting an utterance based on delexicalised dialogue contexts is harder than with lexicalised dialogues.

\begin{table}[!t]
\centering
\def\arraystretch{0.8}
{\footnotesize
\begin{tabularx}{\linewidth}{l YYY}
\toprule
\textbf{Variant}              & \textbf{BLEU}             & \textbf{ROUGE}             & \textbf{METEOR}           \\ \midrule
\multicolumn{4}{c}{\cellcolor[HTML]{EFEFEF}\textbf{Baselines}}                                 \\ \midrule

 Sampling                   & 15.8             & 27.3              & 31.0             \\
Greedy                     & 18.0             & 31.2              & 35.6             \\ \midrule
\multicolumn{4}{c}{\cellcolor[HTML]{EFEFEF}BERT \textbf{Classification-based}}                \\ \midrule
+S2                        & 19.4             & 32.1              & 36.4             \\
+S1:delex+S2               & 19.3             & 32.3              & 36.3        \\
+S1:lex+S2                 & 19.3             & 32.1              & 36.2             \\ \midrule
\multicolumn{4}{c}{\cellcolor[HTML]{EFEFEF}quora-distilroberta \textbf{Classification-based}} \\ \midrule
+S2                        & 19.6             & 32.0              & 36.1             \\
+S1:delex+S2               & \textbf{20.0}             & \textbf{32.8}              & \textbf{36.9}             \\
+S1:lex+S2                 & 19.8             & 32.6              & 36.7             \\ \midrule
\multicolumn{4}{c}{\cellcolor[HTML]{EFEFEF}BERT \textbf{Similarity-based}}                    \\ \midrule
+S2                         & 18.6             & 30.8             & 34.8             \\
+S1:delex+S2               & \textbf{19.6}             & \textbf{32.0}              & \textbf{36.5}             \\
+S1:lex+S2                & 19.1             & 31.7              & 36.0              \\ \bottomrule
\end{tabularx}
}%

\vspace{-1mm}
\caption{Reranking performance with selected model variants based on 20 over-generated responses from MinTL. Full results with other PLMs and variants are available in Table~\ref{tab:plmchoice} in Appendix~\ref{app:multiwoz}.}
\label{tab:performance}
\vspace{-1.5mm}
\end{table}

\subsection{Main Results}

The main results are summarised in Table~\ref{tab:performance}. They suggest that our classification-based reranker yields 20.0 BLEU, outperforming the stronger \textit{Greedy} baseline by 2.0 points. Similar gains are achieved by our similarity-based variant. The comparison of results in Tables~\ref{tab:selection} and \ref{tab:performance} further indicates the inadequacy of standard response selection methods in the reranking task, and the importance of Stage 2 fine-tining: our two-stage reranking framework provides consistent and robust gains over the baselines across the board. Delving deeper into the model performance through ablation experiments, reported in Table~\ref{tab:ablation}, isolates the critical components responsible for the strong performance.

\begin{table}[!t]
\centering
{\footnotesize
\begin{tabularx}{\linewidth}{l X}
\toprule
\textbf{Variant}                                  & \textbf{BLEU }           \\ \midrule
\multicolumn{2}{c}{\cellcolor[HTML]{EFEFEF}\textbf{Classification-based}} \\ \midrule
\textit{quora-distilroberta+S1:delex+S2 }               & 20.0            \\
{ }{ }{ }{ }- self-generated positives         & 13.7 ($\downarrow$6.3)           \\
{ }{ }{ }{ }- multiple positives               & 19.1    ($\downarrow$0.9)           \\
{ }{ }{ }{ }- cross-encoders (+bi-encoders)                  & 15.4 ($\downarrow$4.6)           \\ \midrule
\multicolumn{2}{c}{\cellcolor[HTML]{EFEFEF}\textbf{Similarity-based}}     \\ \midrule
\textit{BERT+S1:delex+S2}                               & 19.6            \\
{ }{ }{ }{ }- self-generated positives         & 15.2 ($\downarrow$4.4)           \\
{ }{ }{ }{ }- multiple positives               & 19.2 ($\downarrow$0.4)           \\ \bottomrule
\end{tabularx}
}%
\vspace{-1mm}
\caption{Ablation experiments on the two best-performing reranking models. We can replace cross-encoders in the classification-based model with the bi-encoder architecture: it encodes contexts and responses separately sharing the encoder's weights; it was trained via the Softmax loss \cite{reimers2019sentence}.}
\label{tab:ablation}
\vspace{-1.5mm}
\end{table}

\vspace{0.6mm}
\noindent \textbf{Self-Generated Positives:} 
Previous work~\cite{krishna2022rankgen} utilises self-generated sentences only to construct negative examples for contrastive learning. Put simply, the model in prior work is trained to select the provided ground truth among self-generated examples. However, for our response reranking task, we need to select the best response from a set where all the items are self-generated. The results suggest that, modeling self-generated responses as positives in S2 (i.e., creating the set $\mathcal{R}_{high}$) is crucial for the reranking effectiveness. The performance degrades considerably for both S2 variants without self-generated positives in S2. The results further suggest that incorporating multiple positive pairs into the same batch yields slight performance gains for both S2 variants. %

\vspace{0.6mm}
\noindent \textbf{Cross-Encoders:} Encoding context-response pairs with cross-encoders instead of using bi-encoders \cite{Humeau2020Poly-encoders,henderson-etal-2020-convert} leads to better performance. Cross-encoders are able to capture finer-grained interactions between the context and the response \cite{geigle-etal-2022-retrieve}, which is pivotal for the response reranking task dealing with subtle variations in the semantically close candidate responses.\footnote{Cross-encoders usually perform better than bi-encoders with the caveat of reduced efficiency~\cite{urbanek2019learning}, but they are typically used exactly in reranking contexts~\cite{geigle-etal-2022-retrieve,Li:2022arxiv} similar to ours.} Further, cross-encoders also enable our similarity-based reranking models.

\begin{table*}[!t]
\centering
\def\arraystretch{0.8}
{\scriptsize
\begin{tabularx}{\textwidth}{l YYY YYY}
\toprule
\multicolumn{1}{c}{\bf S2 Scoring$\downarrow$ / Evaluation$\rightarrow$} & \multicolumn{3}{c}{\bf Classification-based Reranking} & \multicolumn{3}{c}{\bf Similarity-based Reranking} \\
\cmidrule(lr){2-4} \cmidrule(l){5-7}
\multicolumn{1}{c}{}                                  & BLEU            & ROUGE           & METEOR      & BLEU           & ROUGE         & METEOR        \\ \cmidrule(lr){2-4} \cmidrule(l){5-7}
Greedy & {18.0} & {31.2} & {35.6} & {18.0} & {31.2} & {35.6} \\
\hdashline
Similarity                                            & 19.3            & 32.3            & 36.3           & 19.6  & 32.0          & 36.5          \\
\hdashline

BLEU                                                  & {20.3}            &  33.2          &   37.2        &  19.6           & 32.4          & 36.5          \\
ROUGE                                                 & \textbf{20.7}   & \textbf{33.6}   & 37.6         & \textbf{19.8}           & \textbf{32.6} & 36.4          \\
METEOR                                                & 18.2            & 33.4            & \textbf{40.0}  & 17.2               & 32.5          & \textbf{39.1} \\
\bottomrule
\end{tabularx}
}%
\vspace{-1mm}
    \caption{Reranking performance of the \textit{BERT+S1:delex+S2} model variant with different scoring functions in Stage 2 for partitioning responses into the sets $\mathcal{R}_{high}$ and $\mathcal{R}_{low}$. For the models with ROUGE and METEOR as scoring functions, we perform model selection based on the best ROUGE and METEOR performance, respectively.}
\label{tab:metrics}
\vspace{-2mm}
\end{table*}
\captionsetup[subfigure]{oneside,margin={-1cm,2cm},skip=-6pt}
\begin{figure}[!t]
    \centering
    \begin{subfigure}[!ht]{0.431\textwidth}
        \centering
        \includegraphics[width=0.9\linewidth]{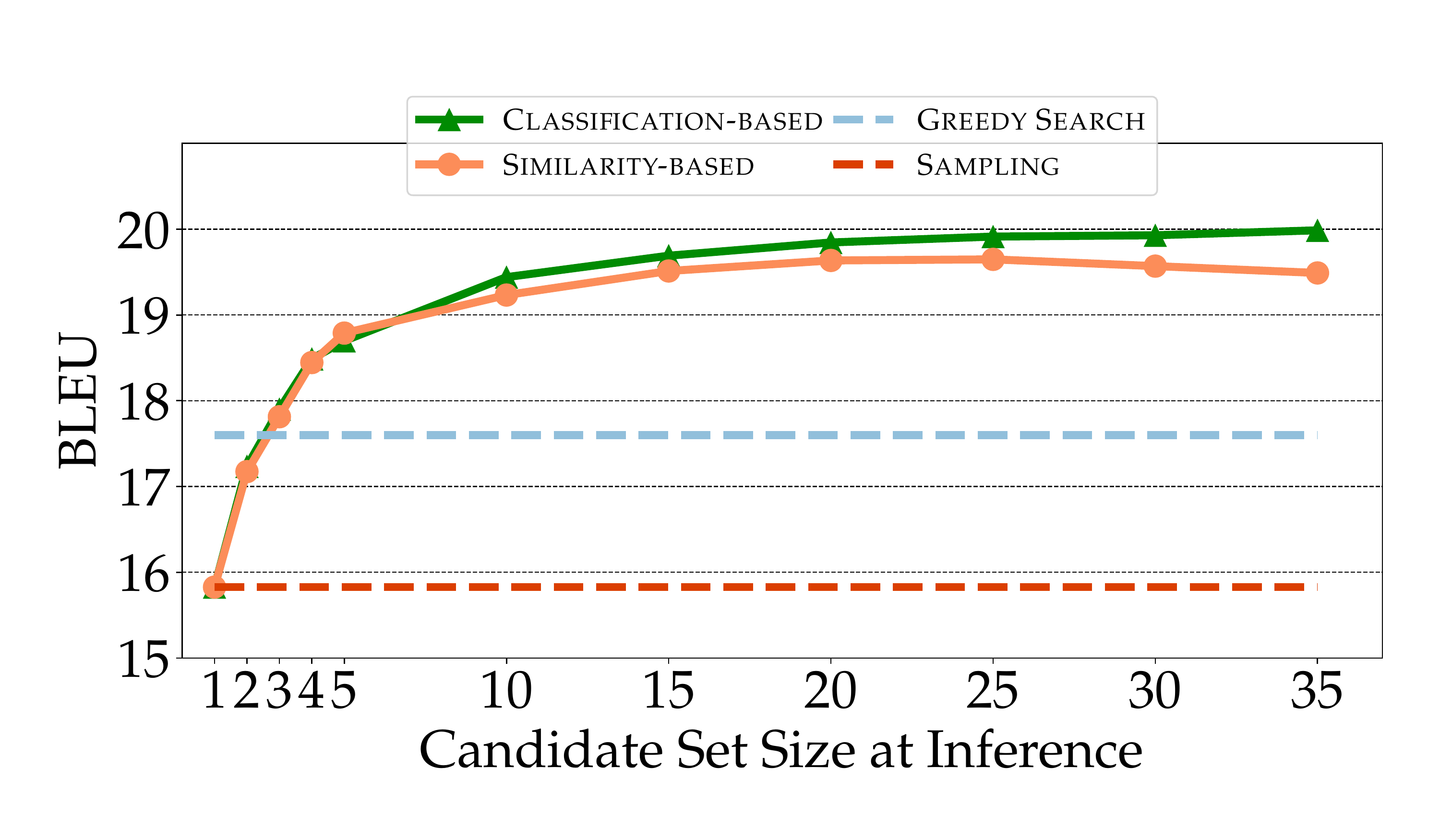}
        \caption{Inference}
        \label{fig:num_overgeneration}
    \end{subfigure}
        \begin{subfigure}[!ht]{0.441\textwidth}
        \centering
        \includegraphics[width=0.9\linewidth]{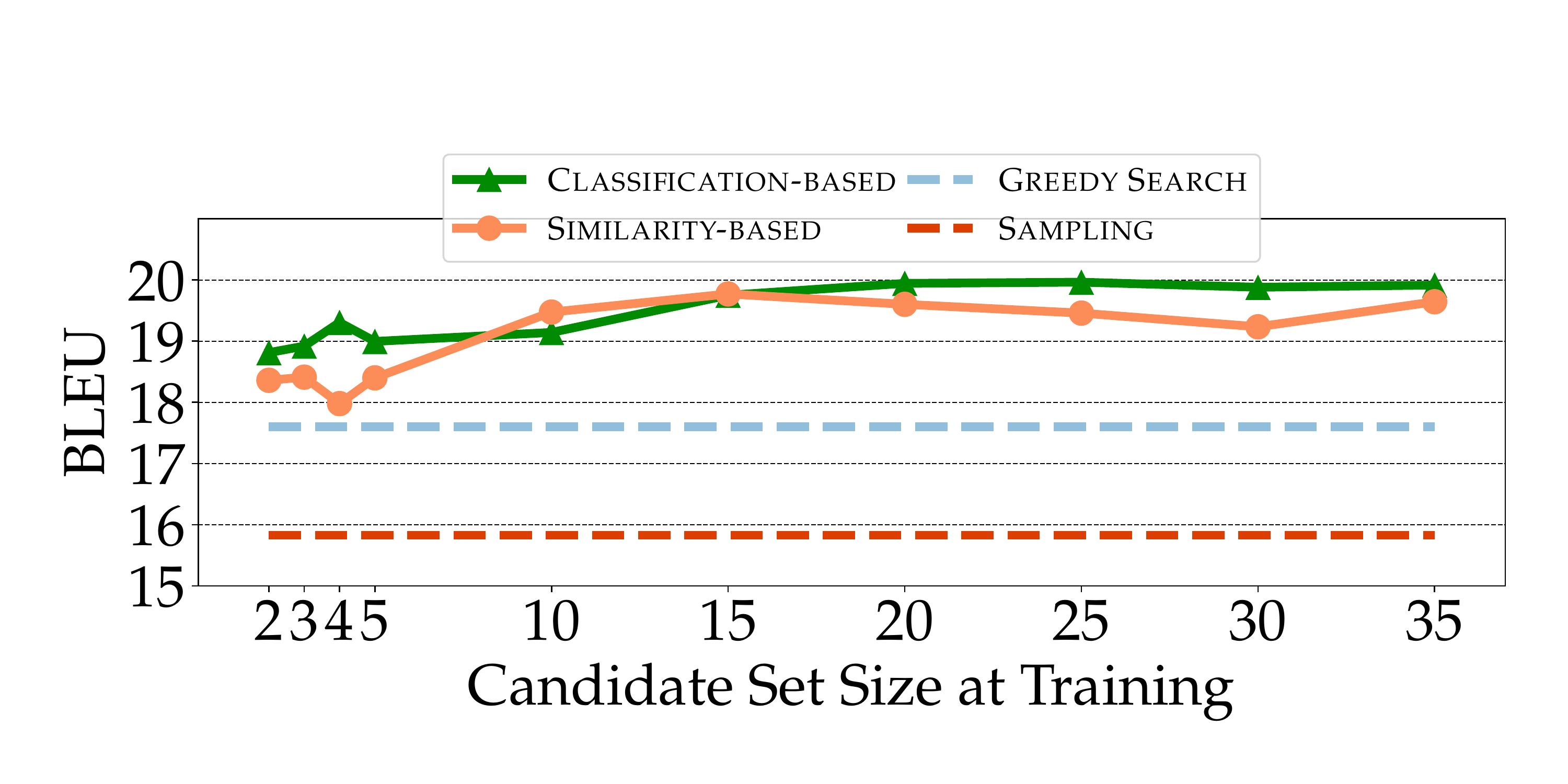}
        \caption{Training}
        \label{fig:num_training}
    \end{subfigure}
    \vspace{-1mm}
    \caption{Impact of the candidate set size for reranking during \textbf{(a)} inference and \textbf{(b)} training on the final BLEU performance. The plots focus on the best-performing model variants from Table~\ref{tab:performance}.}
    \vspace{-1.5mm}
\label{fig:blah}
\end{figure}

\subsection{Further Analysis}
\label{sec:hyperparameter}
We now analyse other important aspects of the proposed reranking framework, running a series of side experiments, with additional analyses of (arguably) lower importance available in Appendix~\ref{app:multiwoz}.

\begin{figure}[!t]
    \centering
    \vspace{-0.3mm}
    \includegraphics[trim={0 0 0 0.8cm}, width=0.8\columnwidth]{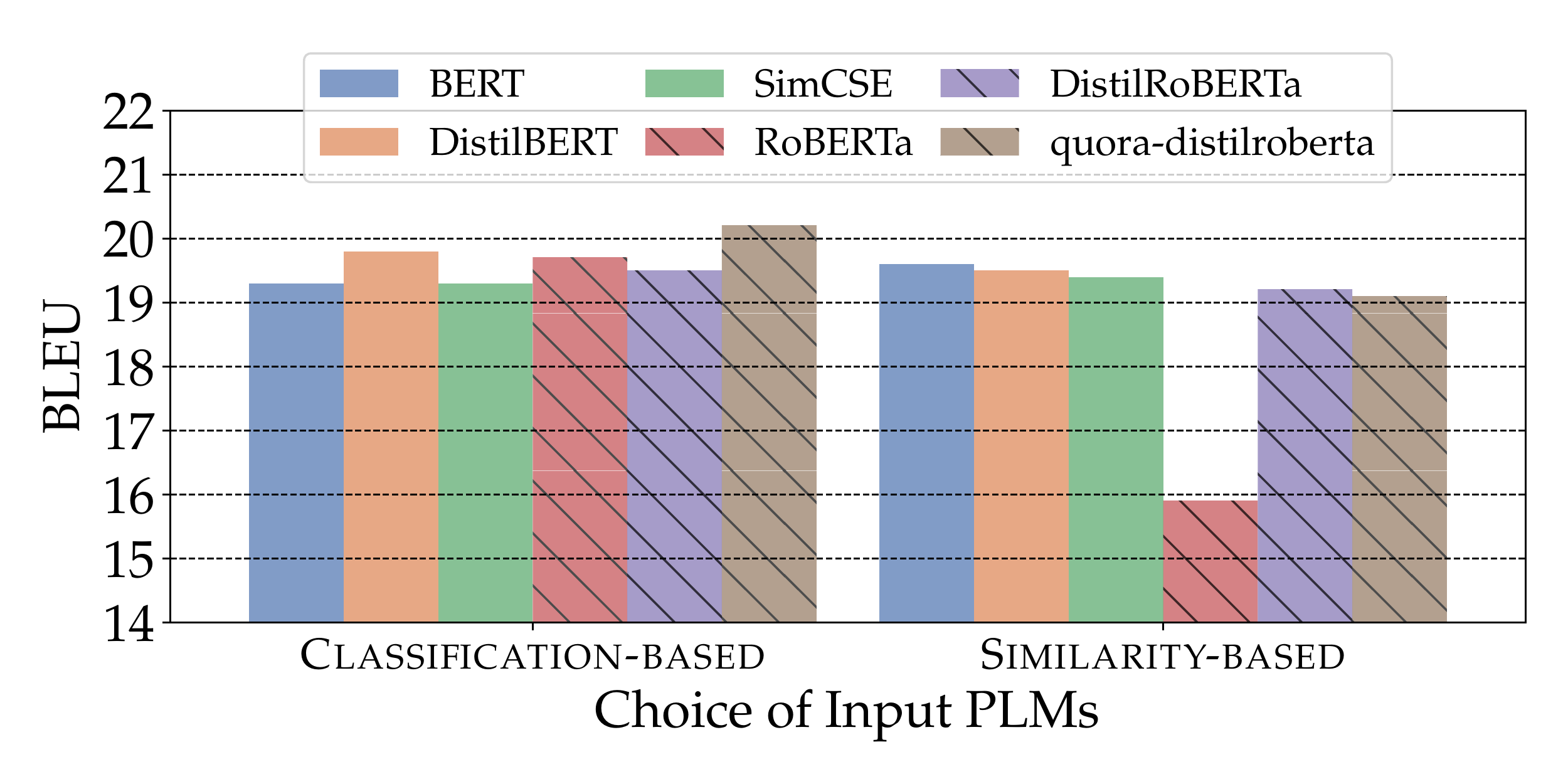}
        \vspace{-2mm}
    \caption{Reranking performance for different input PLMs trained with the entire fine-tuning pipeline (PLM+S1:delex+S2). Full results are in Appendix~\ref{app:multiwoz}.}
    \label{fig:plm_choice}
    \vspace{-2mm}
\end{figure}

\rparagraph{Impact of the Input Encoder}
Figure~\ref{fig:plm_choice} shows the reranking performance with different encoders. Interestingly, the distilled PLMs achieve performance which is on-par with larger models; see also Figures~\ref{fig:plm_choice_rouge}-\ref{fig:plm_choice_meteor} in Appendix~\ref{app:multiwoz}). Further, sentence encoders such as SimCSE and quora-distilroberta do not yield any gains over the other encoders.

\rparagraph{Evaluation Metrics as Scoring Functions}
Table~\ref{tab:metrics} indicates that the gains are consistent across all three automatic evaluation metrics. Moreover, using any of the three metrics as the scoring function $s$ in S2--dividing generated samples into the sets $\mathcal{R}_{high}$ and $\mathcal{R}_{low}$ (see \S\ref{sec:stage2})---yields gains on all the other metrics as well. Naturally, higher scores per each evaluation metric are typically achieved when the same metric is used as the scoring function in S2. However, we note that such a setup might provide artificially inflated metric-specific performance; a better indicator of effectiveness and robustness is the evaluation metric-agnostic model variant used throughout the paper (i.e., the row \textit{Similarity} in Table~\ref{tab:metrics}), which relies on the standard cosine similarity to do partitioning in Stage 2.

\rparagraph{Impact of the Candidate Set Size}
Figure~\ref{fig:num_overgeneration} plots the reranking performance conditioned on a varying number of overgenerated responses during inference.\footnote{The number of overgenerated samples for inference is critical for real-world applications: inference time scales linearly with this number; see more in the Limitations section.} The curves indicate that \textbf{1)} both reranking variants already outperform the Greedy search baseline and achieve SotA performance when the candidate set spans only three candidates; \textbf{2)} after the sharp increase in performance for the sizes 1-10, further increase in the candidate set size offers diminishing returns as performance saturates. In addition, Figure~\ref{fig:num_training} demonstrates that the reranking models outperform the Greedy baseline with only two overgenerated responses for training. Performance also flattens with larger set sizes. %

\begin{figure}[!t]
    \centering
    \includegraphics[trim={0 0 0 2.2cm}, width=0.75\columnwidth]{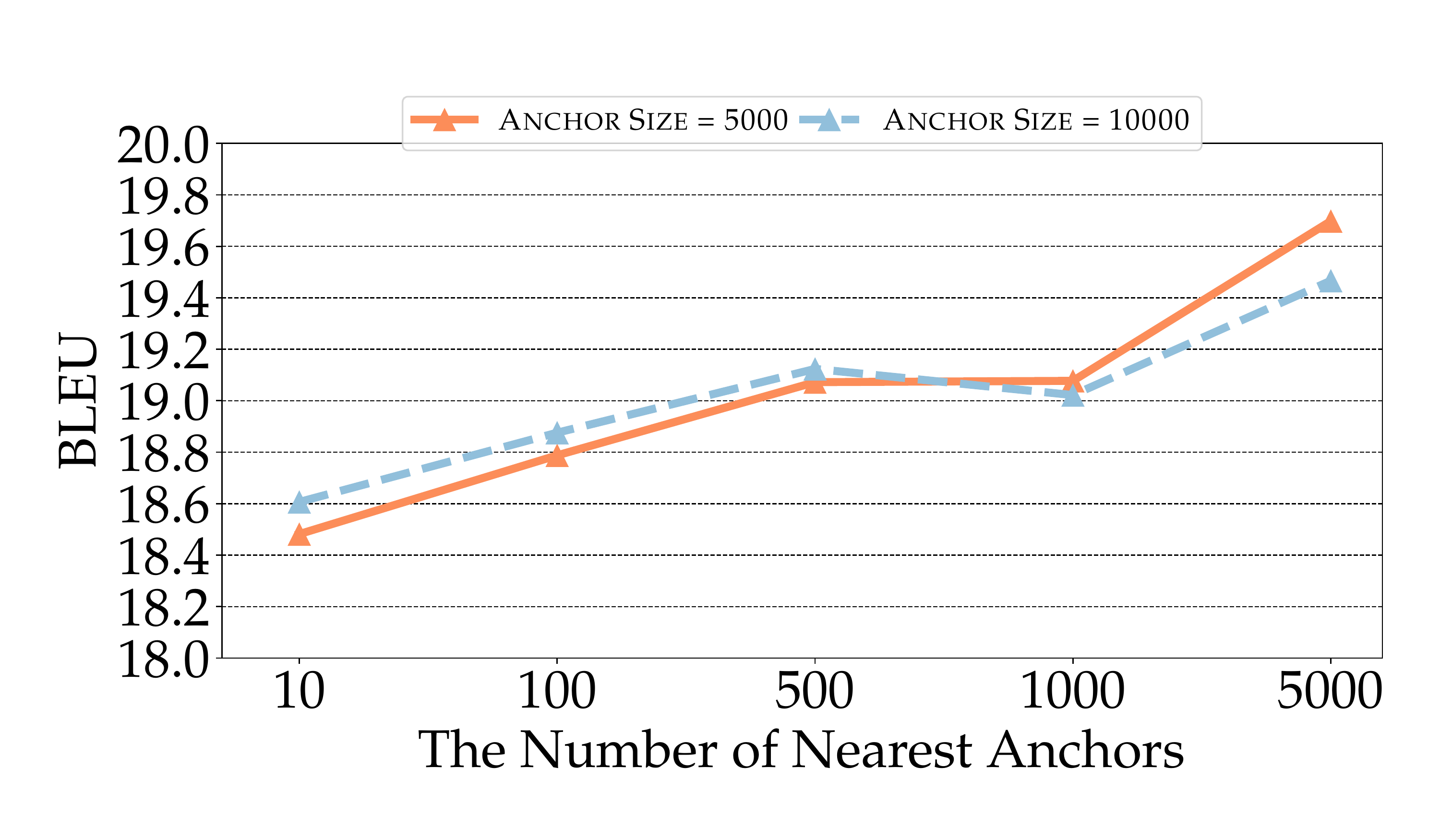}
    \vspace{-2mm}
    \caption{Performance with \textit{BERT+S1:delex+S2:sim} using different values for the total number of anchors and the number of nearest anchors $k$.}
    \label{fig:knn_choice}
    \vspace{-2mm}
\end{figure}

\rparagraph{Similarity-Based Stage 2}
There are two key hyper-parameters: the total number of anchors, and the number of nearest anchors $k$ used to score each test example. Figure~\ref{fig:knn_choice} plots their impact on the final BLEU scores, indicating that the approach is fairly robust to different tested values. Note that the time consumption scales linearly with the anchor pool size (see Table~\ref{tab:time} in Appendix~\ref{sec:experiment_details}).

\rparagraph{Another ToD Dataset}
To test the generalisability of the proposed method, we also run experiments on another ToD dataset: BiToD~\cite{lin2021bitod}. The experimental setup is described in Appendix~\ref{sec:bitod_experiment}, while the results are summarised in Table~\ref{tab:performance_bitod}. Our reranking framework again yields gains over the baselines, verifying its robustness, but the gains are now less pronounced. Delving deeper into the roots of this result, we attribute this to BiToD's data properties combined with the baseline system: mT5 \cite{xue-etal-2021-mt5}, resulting in the lack of syntactic and semantic variability in MinTL's generated outputs on BiToD.  On average, there are only 10.4 unique utterances within the set of 20 overgenerated items, compared to 17.8 for MultiWoZ, with cases where all the 20 generated responses are identical, which leaves meagre or no room for further improvement via reranking.

\begin{table}[!t]
\centering
\def\arraystretch{0.75}
{\footnotesize
\begin{tabularx}{\linewidth}{l YYY}
\toprule
\rowcolor{Gray}
\textbf{Variant}              & BLEU             & ROUGE             & METEOR           \\ \midrule
Beam Search                     & 45.8              & 45.5              & 50.7            \\
Sampling                   & 43.0             & 42.9             & 48.4             \\ %
\midrule
BERT+S1:delex+S2:class                 & 46.3             & 45.9              & 51.4             \\ %
BERT+S1:delex+S2:sim                & 45.9             & 45.4             & 51.3             \\ \bottomrule
\end{tabularx}
}%
\vspace{-1mm}
\caption{Reranking performance based on 20 overgenerated responses on English BiToD.}
\label{tab:performance_bitod}
\vspace{-2mm}
\end{table}

\subsection{Human Evaluation}
\label{ss:human}

\noindent User satisfaction is always the ultimate goal of developing ToD systems~\cite{ji-etal-2022-achieving}. We thus also run additional evaluations with human subjects, with the details on the setup in Appendix~\ref{sec:human_eval_appendix}. We follow suggestions from prior work \cite{fomicheva2021eval4nlp}, and conduct comparative `A/B' tests with 6 participants, each scoring 100 dialogues. Each test item contains a dialogue context and three randomly ordered outputs from greedy search, classification-based reranker, and similarity-based reranker; the human evaluator is tasked with indicating pairwise preferences among the three outputs.
The results indicate:
\begin{enumerate*}[label=(\roman*)]
\item a 49.5\% (0.5\% less) preference for classification-based reranker over the greedy search (0.28 Fleiss' Kappa); 
\item a 57.8\% preference for similarity-based reranker over the greedy search baseline (0.26);
\item a 55.8\% preference for similarity-based reranker over the classification-based reranker (0.16)
\end{enumerate*}.
Overall, we can see a slight preference towards the output of similarity-based rerankers.\footnote{Following \citet{Welleck2020Neural}, we perform 2-sided binomial tests demonstrating the significance of the human preference. See detailed results in Table \ref{tab:human_eval}.} However, we did not observe a strong inter-annotator agreement, as measured by Fleiss' Kappa~\cite{fleiss1971measuring}, as the scoring task is considered highly subjective, and most responses are similar and difficult to distinguish, even for humans (e.g., see an example in Figure~\ref{fig:ui}). This evaluation difficulty again reflects the difficulty of our proposed reranking task, which is a particularly challenging modelling scenario for neural models.

%% file: 06_conclusion.tex
We proposed a novel post-generation reranking method applicable to any end-to-end (E2E) task-oriented dialogue (ToD) system. The reranking is formulated as a two-stage conversational fine-tuning procedure that transforms any input pretrained language model into a specialised in-domain reranker which can operate on the sets of generated responses from the E2E ToD system. Combined with a strong E2E ToD system (MinTL), our reranking models improved E2E dialogue generation performance on standard ToD benchmarks, and achieved new state-of-the-art results on the MultiWOZ leaderboard. Our human evaluation results corroborated the high complexity of E2E ToD evaluation, while again indicating that the proposed rerankers can outperform the standard greedy search baseline. We hope that our work will spark and guide future efforts on using reranking methods for E2E ToD systems.

%% file: 07_limitations.tex
One limitation of the proposed reranking is of practical nature and concerns its dependence on two expensive operations: overgeneration and reranking. In theory, the time complexity of overgeneration scales linearly with the number of outputs, similarly to the beam size for beam search. In practice, we observe that, with the HuggingFace implementation~\cite{wolf2019huggingface}, sampling 10 responses doubles the time consumption compared to sampling a single response. In addition, unlike beam search, this over-sampling can easily be parallelised for real-world applications. Our method improves the baseline even if we only generate three responses during inference (see Figure~\ref{fig:num_overgeneration}). Reranking is less time-demanding, and it takes $\sim$2 minutes on a single GPU (see Appendix~\ref{sec:experiment_details}) for the full MultiWoZ test set with 20 candidates. In future work, we will explore more parameter-efficient methods for over-generation.

Our proposed reranking method is versatile and opens up many further extensions and experimentation beyond the scope and confines of this paper. For instance, we might incorporate the ordering of the self-generated responses and replace the current contrastive loss functions with other recent effective contrastive losses~\cite{zhou2020ladderloss, liu2021simcls}. Furthermore, this paper only explores out-of-box dialogue generation models without further fine-tuning. However, as the comparison of absolute gains on MultiWoZ versus BiToD indicates, increasing \textit{response diversity} leads to a better reranking model and better performance. In future work, we will put more effort on diversifying the set of overgenerated responses in order to harvest more benefits of reranking. 

The current work is also limited only to experiments with the English language, also due to the lack of suitable ToD training data for other languages \cite{Genie:2022survey}. We also plan to extend our model to languages beyond English and other dialogue-generation tasks (e.g. open-domain dialogue generation).

Finally, our work again outlines the complexity and limitations of current evaluation protocols for E2E ToD and ToD in general, as well as the importance of reporting multiple automatic and human-based evaluation metrics.

%% file: xx_appendix.tex
\section{Reranking Variants in Stage 2}
\label{app:stage2_fig}
\begin{figure}[!t]
    \centering
    \includegraphics[width=0.99\linewidth]{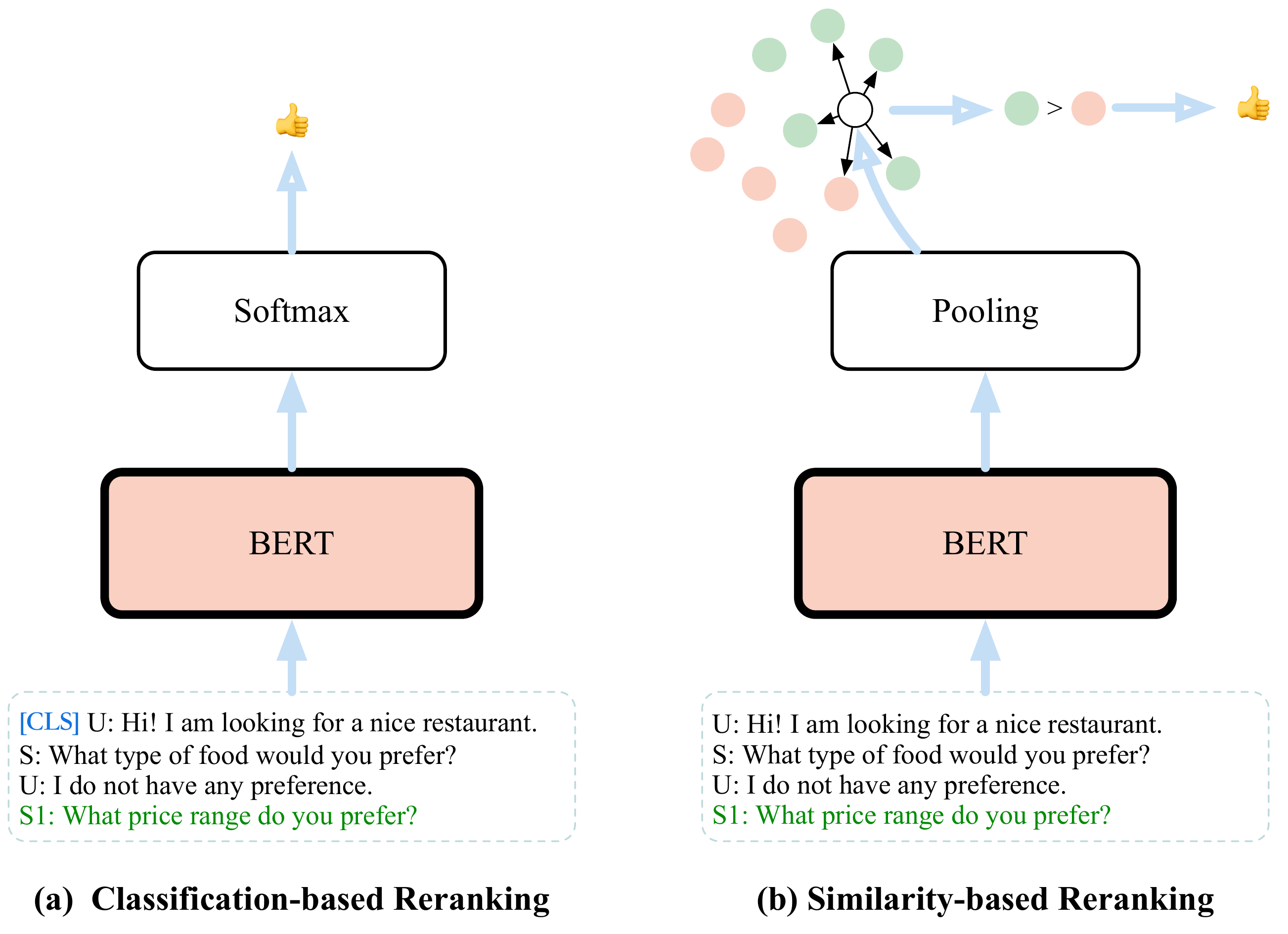}
    \vspace{-1mm}
    \caption{Different reranking variants in Stage 2.}
    \label{fig:fig3}
    \vspace{-2.5mm}
\end{figure}

\textbf{Figure~\ref{fig:fig3}} illustrates the two proposed approaches for Stage 2 fine-tuning, with detailed descriptions available in the main paper (see \S\ref{sec:methodlogy}).

\section{Experimental Details}
\label{sec:experiment_details}

We run all our experiments on a single RTX 24 GiB GPU.

\vspace{1.5mm}
\noindent \textbf{Table \ref{tab:hyperparameters}} lists the search set of our model hyperparameters. Unless mentioned otherwise, all the hyperparameters are set to the default values provided in the SBERT repository. For the classification-based response reranking training, the batch size is 64. For similarity-based reranking training, the batch size is 128. Both batch size values are determined as the maximum values based on our hardware (see above).

\vspace{1.5mm}
\noindent \textbf{Table \ref{tab:input_plms}} lists all the PLMs we used in this work, along with their respective checkpoints in the Huggingface repository.

\vspace{1.5mm}
\noindent \textbf{Table \ref{tab:time}} shows time consumption of our proposed ranking models for fine-tuning and inference. The time consumption is measured based on five independent runs for the BERT-based models on the MultiWoZ dataset.

\vspace{1.5mm}
\noindent \textbf{Impact of Random Initialisation.} 
For our best-performing models, we ran five independent runs with different random seeds. The main finding is that the scores exhibit small-to-negligible variance across different runs. Namely, our best-performing classification-based model achieves the BLEU score of $19.96 \pm 0.16$, and our similarity-based model achieves the BLEU score of $19.59 \pm 0.09$.

\begin{table}[!t]
\centering
{\footnotesize
\begin{tabularx}{\linewidth}{l X}
\toprule
{\bf Hyper-parameter}                                   & {\bf Value}                         \\ \midrule
\multicolumn{2}{c}{\cellcolor[HTML]{EFEFEF}{\bf Stage 1: Response Selection}} \\ \midrule
batch size                              & 64                      \\
context window                          & \{1, 2, 3, 4, 5\}             \\
max sequence length                     & 128                           \\
training epoch                          & \{1, 2, 3, 10\}                             \\
candidate size                              & 20*                            \\ \midrule
\multicolumn{2}{c}{\cellcolor[HTML]{EFEFEF}{\bf Stage 2: Response Reranking}} \\ \midrule
batch size                              & \{64, 128\}                   \\
context window                          & \{1, 2, 3, 4, 5\}             \\
max sequence length                     & 128                   \\
training epoch                          & 5                             \\
BatchAllTripletLoss margin              & 5                             \\ \bottomrule
\end{tabularx}
}%
\caption{Model hyper-parameters. (*)For each dialogue $({c}^{(i)}, {r}^{(i)}) \in \mathcal{D}$, the candidate size for response selection training is 20. In other words, there are 1 positive response ${r}^{(i)}$ and 19 negative responses $\mathcal{R}_{i,-}$. For each ${r}^{(j)} \in \mathcal{R}_{i,-}$ it holds $i \neq j$.}
\label{tab:hyperparameters}
\end{table}
\begin{table}[!t]
\centering
{\footnotesize
\begin{tabularx}{\linewidth}{l X}
\toprule
Hyper-parameter                                   & Value                         \\ \midrule
\multicolumn{2}{c}{\cellcolor[HTML]{EFEFEF}\textbf{Stage 1: Response Selection}} \\ \midrule
batch size                              & 64                    \\
context window                          & 3             \\
max sequence length                     & 128                           \\
training epoch                          & 10                             \\
candidate size                              & 20                            \\ \midrule
\multicolumn{2}{c}{\cellcolor[HTML]{EFEFEF}\textbf{Stage 2: Response Reranking}} \\ \midrule
batch size                              & \{64, 128\}                    \\
context window                          & 3            \\
max sequence length                     & 128                 \\
training epoch                          & 5                             \\
BatchAllTripletLoss mergin              & 5                             \\ \bottomrule
\end{tabularx}
}
\caption{BiToD experiments: model hyperparameters.}
\label{tab:hyperparameters_bitod}
\end{table}

\begin{table}[!t]
\centering
\resizebox{\columnwidth}{!}{

\begin{tabular}{@{}ll@{}}
\toprule
\textbf{Model}               & \textbf{HuggingFace Checkpoint}                     \\ \cmidrule(lr){2-2}
BERT                & bert-base-uncased                          \\
RoBERTa             & roberta-base                               \\
DistilBERT          & distilbert-base-uncased                    \\
DistilRoBERTa       & distilroberta-base                         \\
SimCSE              & princeton-nlp/sup-simcse-bert-base-uncased \\
MiniLM              & sentence-transformers/all-MiniLM-L12-v2    \\
all-mpnet-v2        & sentence-transformers/all-mpnet-base-v2    \\
quora-distilroberta & cross-encoder/quora-distilroberta-base     \\ \bottomrule
\end{tabular}}
\caption{Input PLMs.}
\label{tab:input_plms}
\end{table}

\begin{table}[!t]
\centering
{\footnotesize
\begin{tabularx}{\linewidth}{l X}
\toprule
\textbf{Setup}                                    & \textbf{Time Consumption }            \\ \midrule
\multicolumn{2}{c}{\cellcolor[HTML]{EFEFEF}\textbf{Stage 1: Response Selection}} \\ \midrule
Training per epoch                       & 29:18                        \\
Inference \textit{(full test)      }                        & 1:58                         \\ \midrule
\multicolumn{2}{c}{\cellcolor[HTML]{EFEFEF}\textbf{Stage 2: Response Reranking}} \\ \midrule
Training per epoch                       & 27:10                        \\
Inference with 5,000 anchors \textit{(full test) }             & 2:20                         \\
Inference with 10,000 anchors \textit{(full test)  }     & 4:52                         \\ \bottomrule
\end{tabularx}
}%
\caption{Time consumption of our proposed methods. It was computed as an average of 5 runs on a machine with a 16-core vCPU and a single RTX 24 GiB GPU.}
\label{tab:time}
\end{table}

\section{Additional Results on MultiWoZ}
\label{app:multiwoz}

\begin{figure}[!t]
    \centering
    \includegraphics[width=\columnwidth]{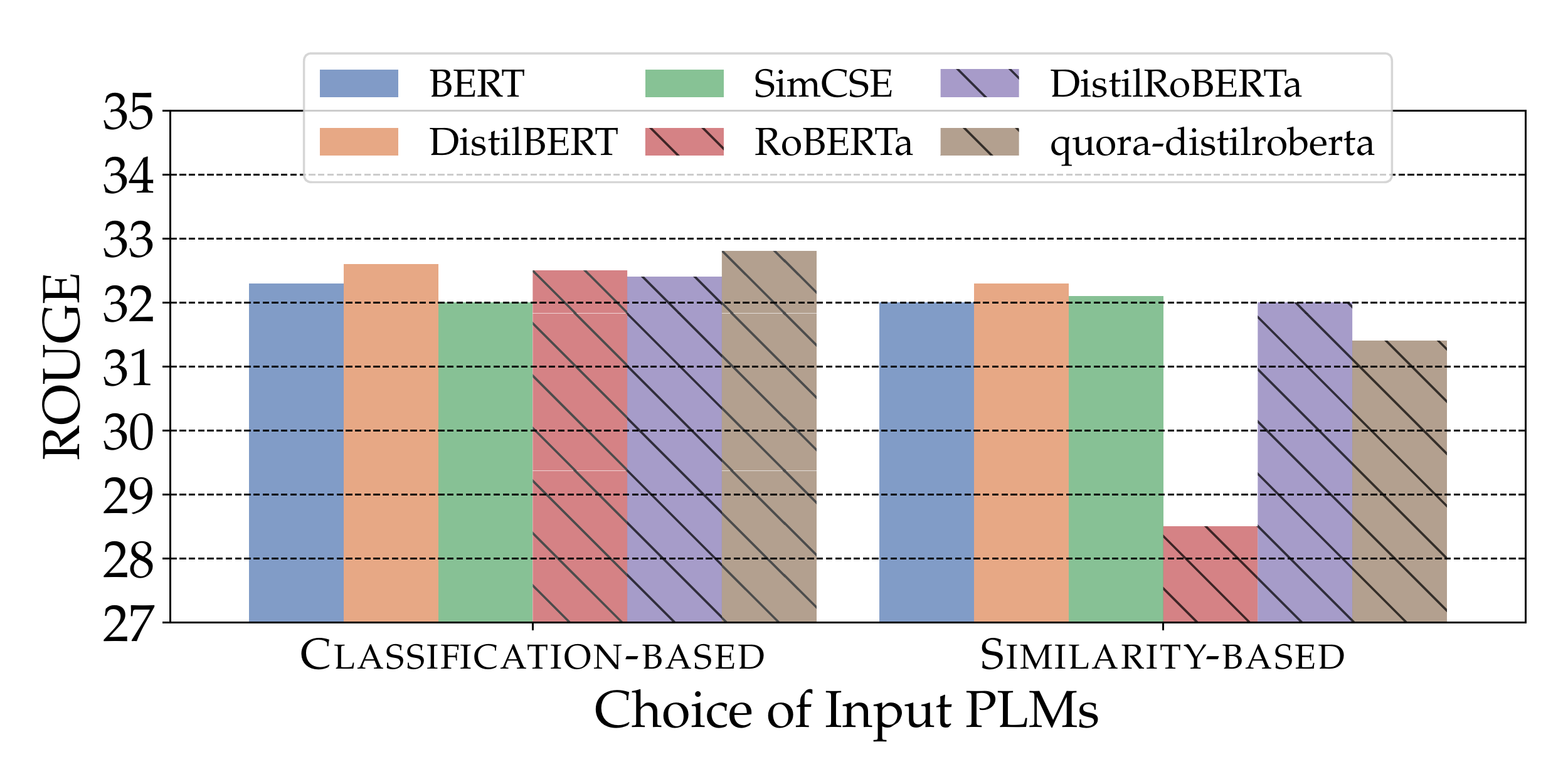}
    \caption{Reranking performance (ROUGE scores) for different input PLMs trained with the entire fine-tuning pipeline (PLM+S1:delex+S2)..}
    \label{fig:plm_choice_rouge}
\end{figure}

\begin{figure}[!t]
    \centering
    \includegraphics[width=\columnwidth]{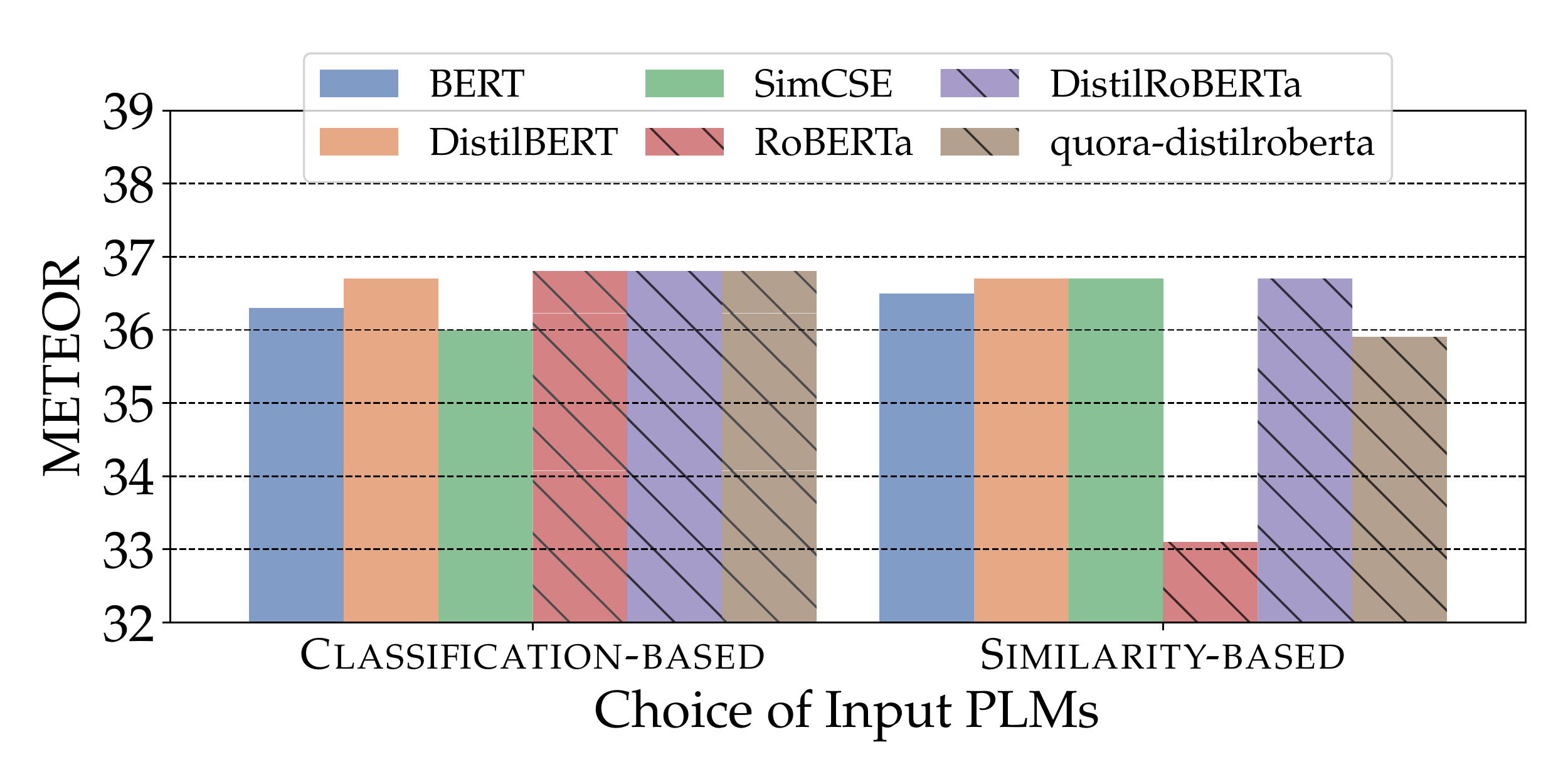}
    \caption{Reranking performance (METEOR scores) for different input PLMs trained with the entire fine-tuning pipeline (PLM+S1:delex+S2).}
    \label{fig:plm_choice_meteor}
\end{figure}

To solidify our findings in this paper, we list additional experimental results which offer further empirical support for our main claims:

\vspace{1.5mm}
\noindent \textbf{Figure \ref{fig:plm_choice_rouge}} and \textbf{Figure \ref{fig:plm_choice_meteor}} demonstrate the reranking performance with different input PLMs, measured by the ROUGE and METEOR score, respectively (see also \S\ref{sec:hyperparameter}).

\vspace{1.5mm}
\noindent \textbf{Table \ref{tab:plmchoice}} provides the results with different input PLMs in our comparison with the full fine-tuning pipeline. From this table, sentence encoders do not provide advantages over PLMs. Table \ref{tab:plmchoice} can been seen as an expanded version of Figure \ref{fig:plm_choice}, Figure \ref{fig:plm_choice_rouge}, and Figure \ref{fig:plm_choice_meteor}.

\vspace{1.5mm}
\noindent \textbf{Table~\ref{tab:context_window}} provides the results with the \textit{BERT+S1:delex+S2} model variant with varying dialogue history/context size. Both classification-based and similarity-based rerankers utilise the historical dialogue context and require at least 2 preceding historical utterances to be effective.
However, there is no discernible correlation between the reranking performance and the dialogue context size further beyond. In other words, by increasing or decreasing the number of the input historical utterances, the reranking performance does not catastrophically degrade, when more than two historical utterances are available.

\begin{table*}[!t]
\centering
\resizebox{\textwidth}{!}{%
\begin{tabular}{@{}lclccclccc@{}}
\toprule
\multicolumn{1}{c}{}                                & \textbf{Response Selection} &  & \multicolumn{3}{c}{\bf Classification-based Reranking} &  & \multicolumn{3}{c}{\bf Similarity-based Reranking} \\ \cmidrule(lr){2-2} \cmidrule(lr){4-6} \cmidrule(l){8-10} 
\multicolumn{1}{c}{\multirow{-2}{*}{\bf Variant}} & R@1                &  & BLEU           & ROUGE           & METEOR          &  & BLEU          & ROUGE         & METEOR         \\ \midrule
\multicolumn{10}{c}{\cellcolor[HTML]{EFEFEF} {\bf Random Baseline}}                                                                                                                         \\ \midrule
Random Sampling                                     & 5.0                &  & 15.8           & 27.3            & 31.0            &  & 15.8          & 27.3          & 31.0           \\ \midrule
\multicolumn{10}{c}{\cellcolor[HTML]{EFEFEF} {\bf Sentence Encoders}}                                                                                                                      \\ \midrule
BERT                                                & n/a                &  & n/a            & n/a             & n/a             &  & 17.0          & 29.4          & 33.6           \\
SimCSE                                              & n/a                &  & n/a            & n/a             & n/a             &  & 16.7          & 29.0          & 33.2           \\ 
quora-distilroberta                                 & n/a                &  & n/a            & n/a             & n/a             &  & 15.9          & 27.7          & 32.0           \\
all-mpnet                                           & n/a                &  & n/a            & n/a             & n/a             &  & 16.0          & 27.6          & 31.8           \\ \midrule
\multicolumn{10}{c}{\cellcolor[HTML]{EFEFEF}{\bf Response Selection Models}}                                                                                                               \\ \midrule
BERT+S1:delex                                       & 51.0               &  & 16.2           & 29.0            & 34.2            &  & 16.7          & 39.3          & 33.8           \\
DistilRoBERTa+S1:delex                              & 48.0               &  & 16.1           & 38.9            & 33.7            &  & 16.6          & 29.0          & 33.4           \\
quora-distilroberta+S1:delex                        & 50.5               &  & 16.3           & 28.8            & 33.6            &  & 16.6          & 29.1          & 33.9           \\
BERT+S1:lex                                         & 77.2               &  & 14.5           & 28.8            & 33.2            &  & 17.1          & 29.7          & 34.3           \\
DistilRoBERTa+S1:lex                                & 74.4               &  & 15.0           & 27.7            & 32.4            &  & 16.6          & 29.6          & 34.5           \\
quora-distilroberta+S1:lex                          & 74.9               &  & 15.2           & 27.9            & 32.8            &  & 16.2          & 28.6          & 33.2           \\ \bottomrule
\end{tabular}%
}
\caption{Response selection and response reranking performance on the MultiWoZ test set with standard response selection models and out-of-box sentence encoders.  n/a = non-applicable. Selected results from this table are in Table~\ref{tab:selection} in the main paper.}
\label{tab:selection_old}
\end{table*}

\begin{table*}[!t]
\centering
{\footnotesize
\begin{tabular}{@{}llllllll@{}}
\toprule
\multicolumn{1}{c}{}                                & \multicolumn{3}{c}{\bf Classification-based Reranking} &  & \multicolumn{3}{c}{\bf Similarity-based Reranking} \\ \cmidrule(lr){2-4} \cmidrule(l){6-8} 
\multicolumn{1}{c}{\multirow{-2}{*}{\bf Variant}} & BLEU           & ROUGE           & METEOR          &  & BLEU          & ROUGE         & METEOR         \\ \midrule
\multicolumn{8}{c}{\cellcolor[HTML]{EFEFEF}\textbf{PLMs}}                                                                                                      \\ \midrule
BERT+S1:delex+S2                                    & 19.3           & 32.3            & 36.3            &  & 19.6          & 32.0          & 36.5           \\
DistilBERT+S1:delex+S2                              & 19.8           & 32.6            & 36.7            &  & 19.5          &  32.3         & 36.7            \\
RoBERTa+S1:delex+S2                                 & 19.7           & 32.5            & 36.8            &  &  15.9         &  28.5         &  33.1          \\

DistilRoBERTa+S1:delex+S2                           & 19.5           & 32.4            & 36.7            &  &  19.2         &   32.0        & 36.8           \\
BERT+S1:lex+S2                                    & 19.3           & 32.1 $\downarrow$            & 36.2 $\downarrow$            &  & 19.1 $\downarrow$          & 31.7 $\downarrow$          & 36.0 $\downarrow$           \\
DistilBERT+S1:lex+S2                              & 19.7 $\downarrow$           & 32.4 $\downarrow$            & 36.5 $\downarrow$            &  & 19.8          &  32.1 $\downarrow$         & 36.5 $\downarrow$            \\
RoBERTa+S1:lex+S2                                 & 19.9            & 32.7           & 36.7 $\downarrow$           &  &  19.0         &  31.7         &  36.3          \\
DistilRoBERTa+S1:lex+S2                           & 19.4 $\downarrow$           & 32.1 $\downarrow$            & 36.3 $\downarrow$            &  &  18.6 $\downarrow$         &   31.8 $\downarrow$        & 37.1        \\
\midrule
\multicolumn{8}{c}{\cellcolor[HTML]{EFEFEF}\textbf{Sentence Encoders}}                                                                                                   \\ \midrule
MiniLM+S1:delex+S2                                  & 19.9           & 32.4            & 36.5            &  & 19.0           & 32.0      & 37.1           \\ 
all-mpnet+S1:delex+S2                               & 19.7           & 32.4            & 36.5            &  & 18.9           & 32.1      & 36.9           \\ 
SimCSE+S1:delex+S2                                  & 19.3           & 32.0            & 36.0            &  & 19.4           & 32.1      & 36.7           \\ 
quora-distilroberta+S1:delex+S2                     & 20.0           & 32.8           & 36.9            &  &  19.1         &  31.4         &   35.9         \\ 
MiniLM+S1:lex+S2                                  & 19.3 $\downarrow$           & 31.9 $\downarrow$            & 35.9 $\downarrow$            &  & 18.9 $\downarrow$           & 31.4 $\downarrow$      & 35.8 $\downarrow$           \\ 
all-mpnet+S1:lex+S2                               & 19.6 $\downarrow$           & 32.3 $\downarrow$            & 36.5           &  & 18.8 $\downarrow$          & 31.5 $\downarrow$      & 35.8 $\downarrow$           \\ 
SimCSE+S1:lex+S2                                  & 19.5           & 32.4            & 36.5            &  & 19.0 $\downarrow$          & 31.4 $\downarrow$      & 35.7 $\downarrow$           \\ 
quora-distilroberta+S1:lex+S2                     & 19.8 $\downarrow$           & 32.6 $\downarrow$           & 36.7 $\downarrow$            &  &  18.3 $\downarrow$         &  31.3 $\downarrow$         &   35.9        \\
\bottomrule
\end{tabular}%
}%
\caption{Response reranking models trained with the full fine-tuning pipeline with different input PLMs. $\downarrow$ denotes a lower performance compared to the counterpart model trained with delexicalised dialogues.}
\label{tab:plmchoice}
\end{table*}

\begin{table*}[!t]
\centering
{\footnotesize
\begin{tabularx}{\textwidth}{l XXX XXX}
\toprule
\multicolumn{1}{c}{\multirow{2}{*}{\bf Context Size}} & \multicolumn{3}{c}{\bf Classification-based Reranking} &  \multicolumn{3}{c}{\bf Similarity-based Reranking} \\ \cmidrule(lr){2-4} \cmidrule(l){5-7} 
\multicolumn{1}{c}{}                              & BLEU           & ROUGE           & METEOR          & BLEU          & ROUGE         & METEOR         \\ \midrule
1                                                 & 16.7           & 29.8            & 34.5            & 16.7          & 29.7          & 34.5           \\
2                                                 & 19.2           & 32.4            & 36.7            & 18.8          & 32.3          & 37.2           \\
3                                                 & 19.3           & 32.3            & 36.3             & 19.6          & 32.0          & 36.5           \\
4                                                 & 19.5           & 32.4            & 36.6             & 19.2          & 31.5          & 35.7           \\
5                                                 & 19.4           & 32.3            & 36.2            & 19.6          & 32.1          & 36.2           \\ \bottomrule
\end{tabularx}
}%

\caption{Reranking performance of the \textit{BERT+S1:delex+S2} variant with different context window sizes. The context window size is the number of historical dialogue utterances for the reranking models. We choose the same window size for fine-tuning at Stage 1, and note that the default context window size for the MinTL model is 3.}
\label{tab:context_window}
\end{table*}

\section{BiToD: Experimental Setup}
\label{sec:bitod_experiment}

BiToD~\cite{lin2021bitod} is a bilingual (English and Chinese) multi-domain dataset for end-to-end task-oriented dialogue modeling. For our experiments, we only use the English partition of the whole dataset, which contains 2,952/295/442 dialogues for training/validation/testing. BiToD covers five domains: attraction, hotel, restaurant, weather, and metro. We use the provided baseline system implemented with the mT5 model~\cite{xue-etal-2021-mt5}. We follow the default training script, only reducing the batch size due to hardware constraints (from 8 to 2).

We follow the setup from \S\ref{sec:setup} for our reranking experiments. Following the baseline system, we only train our reranking model with lexicalised dialogues. \textbf{Table \ref{tab:hyperparameters_bitod}} lists the chosen and search set of our model hyperparameters. We searched the $k \in \{10, 100, 500, 1000, 5000\}$ for KNN regression and the number of anchors $n \in \{5000, 10000\}$. In addition, for BiToD we remove the downsampling step in Stage 2, which led to better empirical results.

\section{Human Evaluation}
\label{sec:human_eval_appendix}

\begin{figure*}[!t]
    \centering
    \includegraphics[width=\textwidth]{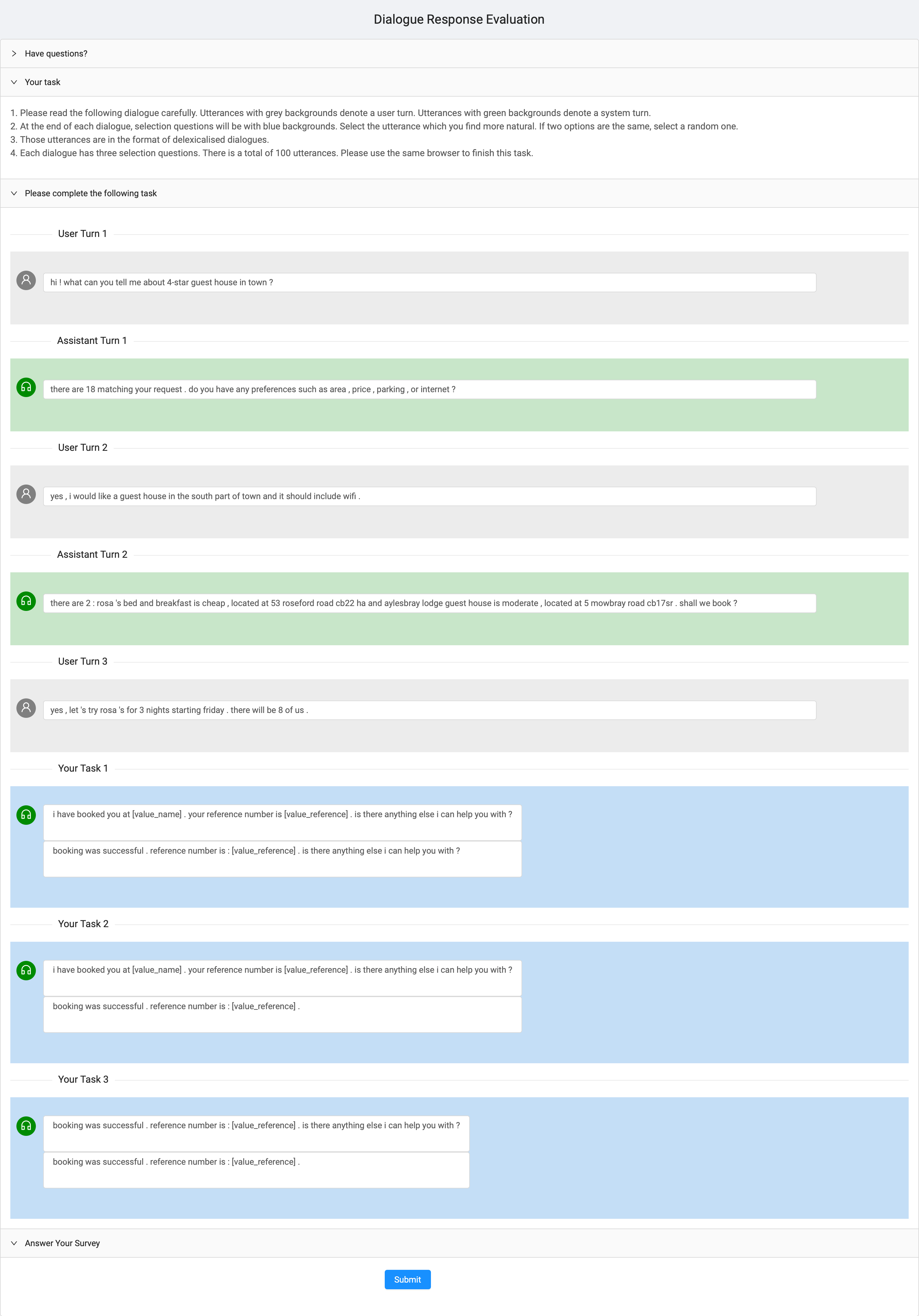}
    \caption{Human evaluation: an example task for human-based A/B testing.}
    \label{fig:ui}
\end{figure*}

We invited six human participants to join our human evaluation experiments. Each annotator answered 100 questions indicating preferences over different model outputs. Those questions are formalised as A/B tests. Recently, comparative evaluation measurements have been shown to be more robust in human evaluation~\cite{fomicheva2021eval4nlp}. As demonstrated by \textbf{Figure \ref{fig:ui}}, given a dialogue context, the human evaluators are instructed to select better responses from pairs of responses generated from different models. We have compared among outputs from our classification-based models (quora-distilroberta), similarity-based models (BERT), and the greedy search baseline (best performing models in Table~\ref{tab:performance}). The ordering of the tasks and models is completely randomised.

\vspace{1.5mm}
\noindent \textbf{Table~\ref{tab:human_eval}} shows full evaluation results. In addition, \textbf{Table~\ref{tab:example}} shows some example dialogues in the MultiWoZ test set, where the similarity-based reranked response is strongly preferred.

\begin{table*}[!t]
\centering
{\footnotesize
\begin{tabularx}{\textwidth}{l XXXXX l}
\toprule
\textbf{Method A vs Method B}                     & \# of A & \# of B & \% of A & \% of B & Total \#  & Fleiss' Kappa\\
\cmidrule(lr){2-7}
Classification-based vs Greedy           & 297     & 303     & 49.5    & 50.5    & 600                    & 0.28        \\
Similarity-based vs Greedy               & 347     & 253     & *57.8    & *42.2    & 600                    & 0.26        \\
Similarity-based vs Classification-based & 335     & 265     & *55.8    & *44.2    & 600                    & 0.16        \\ \bottomrule
\end{tabularx}%
}%
\caption{Number (\#) and percentage (\%) of preferred responses during human A/B testing. Fleiss' Kappa is listed as a measure of inter-annotator agreement. * denotes statistical significance (2-sided binomial test, p < .05).}
\label{tab:human_eval}
\end{table*}

\begin{table*}[!t]
\centering
\resizebox{\textwidth}{!}{%
\begin{tabular}{@{}ll@{}}
\toprule
\multicolumn{2}{c}{\cellcolor[HTML]{EFEFEF}\textbf{Example 1}}                                                                                                                                                                                                                                                                                                                                                                       \\ \midrule
History              & \begin{tabular}[c]{@{}l@{}}......\\ S: i would recommend christ 's college . it is free to visit .\\ U:     could i have their postcode please ?\\ S: christ 's college 's post code is cb23bu\\ U:     thank you , that is all the information i need\end{tabular}                                                                                                                                  \\ \midrule
Ground Truth         & S: you are welcome , please contact us if we can help you with anything else .                                                                                                                                                                                                                                                                                                                       \\ \midrule
Greedy               & S: the postcode is {[}value\_postcode{]} . is there anything else i can help you with ?                                                                                                                                                                                                                                                                                                              \\ \midrule
Classification-based & S: you are welcome . have a great day!                                                                                                                                                                                                                                                                                                                                                               \\ \midrule
Similarity-based     & S: ok, you are welcome . have a great day!                                                                                                                                                                                                                                                                                                                                                           \\ \midrule
\multicolumn{2}{c}{\cellcolor[HTML]{EFEFEF}\textbf{Example 2}}                                                                                                                                                                                                                                                                                                                                                                       \\ \midrule
History              & \begin{tabular}[c]{@{}l@{}}U:     i am planning a trip in cambridge\\ S: great ! is there a certain restaurant you are looking for ?\\ U:     moderate price in the east . and i need the phone number , address , and food type\\ S: i have the curry prince which is indian food . would you like that 1 ?\\ U:     yes . that would be fine . what is the address and phone number ?\end{tabular} \\ \midrule
Ground Truth         & S: the address is {[}value\_address{]} and the phone number is {[}value\_phone{]} . would you like me to make a reservation for you ?                                                                                                                                                                                                                                                                \\ \midrule
Greedy               & S: {[}value\_name{]} is an {[}value\_food{]} restaurant in the {[}value\_area{]} . their address is {[}value\_address{]} . their phone number is {[}value\_phone{]} .                                                                                                                                                                                                                                \\ \midrule
Classification-based & S: the address is {[}value\_address{]} and the phone number is {[}value\_phone{]} .                                                                                                                                                                                                                                                                                                                  \\ \midrule
Similarity-based     & S: the address is {[}value\_address{]}, and the phone number is {[}value\_phone{]} .                                                                                                                                                                                                                                                                                                                 \\ \midrule
\multicolumn{2}{c}{\cellcolor[HTML]{EFEFEF}\textbf{Example 3}}                                                                                                                                                                                                                                                                                                                                                                       \\ \midrule
History              & \begin{tabular}[c]{@{}l@{}}......\\ U: thank you . can you also book a taxi for me ?\\ S: sure , i need to know destination and where you need picked up . i also need the time and number of people .\\ U: i also need a taxi from the church so i can get to the restaurant on time ."\end{tabular}                                                                                                   \\ \midrule
Ground Truth         & S: i will work on getting that booked for you .                                                                                                                                                                                                                                                                                                                                                      \\ \midrule
Greedy               & S: i can help with that . where will you be departing from ?                                                                                                                                                                                                                                                                                                                                         \\ \midrule
Classification-based & S: your taxi has been booked . it will be a {[}value\_car{]} and the contact number is {[}value\_phone{]} . is there anything else i can help you with ?                                                                                                                                                                                                                                             \\ \midrule
Similarity-based     & S: your taxi is booked . it will be a {[}value\_car{]} and the contact number is {[}value\_phone{]} .                                                                                                                                                                                                                                                                                                \\ \midrule
\multicolumn{2}{c}{\cellcolor[HTML]{EFEFEF}\textbf{Example 4}}                                                                                                                                                                                                                                                                                                                                                                       \\ \midrule
History              & \begin{tabular}[c]{@{}l@{}}......\\ S: tr8259 will arrive in cambridge at 10:23 . would you like me to book a ticket for you on that train ?\\ U: not yet . first i would like to get price and departure time .\\ S: it will depart at 07:40 and it will cost 60.08 pounds . do you need help with anything else ?\\ U: and i need a place to stay\end{tabular}                                     \\ \midrule
Ground Truth         & S: okay there are {[}value\_choice{]} options . do you have a price preference or area                                                                                                                                                                                                                                                                                                               \\ \midrule
Greedy               & S: i am sorry, but i don't have any {[}value\_type{]} that meet your criteria . would you like to try a different price range or area ?                                                                                                                                                                                                                                                              \\ \midrule
Classification-based & S: okay, what area would you like to stay in ?                                                                                                                                                                                                                                                                                                                                                       \\ \midrule
Similarity-based     & S: i can help you with that . what area of town are you wanting to stay in ?                                                                                                                                                                                                                                                                                                                         \\ \bottomrule
\end{tabular}%
}
\caption{Selected example dialogues and corresponding responses where all five evaluators prefer the similarity-based reranked output compared to the greedy search output. S: system, U: user.}
\label{tab:example}
\end{table*}